\documentclass{article}

\usepackage[nonatbib,final]{neurips_2025}

\usepackage[utf8]{inputenc} % allow utf-8 input
\usepackage[T1]{fontenc}    % use 8-bit T1 fonts
\usepackage{hyperref}       % hyperlinks
\usepackage{url}            % simple URL typesetting
\usepackage{booktabs}       % professional-quality tables
\usepackage{amsfonts}       % blackboard math symbols
\usepackage{nicefrac}       % compact symbols for 1/2, etc.
\usepackage{microtype}      % microtypography
\usepackage{xcolor}         % colors

\usepackage{algorithm}
\usepackage[noend]{algorithmic}

\usepackage{graphicx}  % resize box
\usepackage{bbding}
\usepackage{xcolor}
\usepackage{multirow}
\usepackage{wrapfig}

\newcommand{\meansd}[2]{#1 $\pm$ #2}  % mean and standard deviation

\usepackage{colortbl}
\definecolor{Gray}{gray}{0.90}
\newcolumntype{g}{>{\columncolor{Gray}}c}

\usepackage{amsmath,amsfonts,bm}

\def\eqref#1{equation~\ref{#1}}
\def\1{\bm{1}}

\def\vzero{{\bm{0}}}
\def\vone{{\bm{1}}}
\def\vmu{{\bm{\mu}}}

\def\vg{{\bm{g}}}

\def\vt{{\bm{t}}}

\def\vv{{\bm{v}}}
\def\vw{{\bm{w}}}

\def\vy{{\bm{y}}}
\def\vz{{\bm{z}}}

\def\mT{{\bm{T}}}

\DeclareMathAlphabet{\mathsfit}{\encodingdefault}{\sfdefault}{m}{sl}
\SetMathAlphabet{\mathsfit}{bold}{\encodingdefault}{\sfdefault}{bx}{n}

\def\gV{{\mathcal{V}}}

\def\sB{{\mathbb{B}}}

\newcommand{\E}{\mathbb{E}}

\newcommand{\R}{\mathbb{R}}

\newcommand{\Var}{\mathrm{Var}}

\newcommand{\Cov}{\mathrm{Cov}}
\DeclareMathOperator*{\argmax}{arg\,max}

\usepackage{amsthm}
\usepackage{thm-restate}

\theoremstyle{plain}
\newtheorem{theorem}{Theorem}[section]

\newtheorem{lemma}[theorem]{Lemma}

\theoremstyle{definition}

\theoremstyle{remark}
\newtheorem{remark}[theorem]{Remark}

\newcommand{\algname}{\texttt{Mint}} % algorithm name

\newcommand{\gVGTtotal}{\gV^{\text{GT}}_{\text{total}}}
\newcommand{\gVGTinter}{\gV^{\text{GT}}_{\text{inter}}}
\newcommand{\gVGTintra}{\gV^{\text{GT}}_{\text{intra}}}

\newcommand{\gVPLtotal}{\gV^{\text{PL}}_{\text{total}}}
\newcommand{\gVPLinter}{\gV^{\text{PL}}_{\text{inter}}}
\newcommand{\gVPLintra}{\gV^{\text{PL}}_{\text{intra}}}

\newcommand{\vdelta}{\boldsymbol{\delta}}
\newcommand{\vsigma}{\boldsymbol{\sigma}}

\newcommand{\vvcls}{\vv^{\text{cls}}}
\newcommand{\vvirr}{\vv^{\text{irr}}}
\newcommand{\vvshift}{\vv^{\text{shift}}}
\newcommand{\vvnoise}{\vv^{\text{noise}}}

\newcommand{\vwcls}{\vw^{\text{cls}}}
\newcommand{\vwirr}{\vw^{\text{irr}}}
\newcommand{\vwshift}{\vw^{\text{shift}}}
\newcommand{\vwnoise}{\vw^{\text{noise}}}

\newcommand{\dcls}{d_{\text{cls}}}
\newcommand{\dirr}{d_{\text{irr}}}
\newcommand{\dshift}{d_{\text{shift}}}
\newcommand{\dnoise}{d_{\text{noise}}}

\newcommand{\sigmayhy}{\sigma_{\hat{y}y}}
\newcommand{\vsigmairr}{\vsigma_{\text{irr}}}
\newcommand{\vsigmanoise}{\vsigma_{\text{noise}}}

\newcommand{\normalize}{\mathrm{normalize}}

\newcommand{\Kprior}{K_{\text{prior}}}

\title{{\algname}: A Simple Test-Time Adaptation of Vision-\\ Language Models against Common Corruptions}

\author{%
  Wenxuan Bao$^{1*}$, \quad Ruxi Deng$^{1*}$, \quad Jingrui He$^1$ \\
  $^1$University of Illinois Urbana-Champaign \\
  \texttt{\{wbao4,ruxid2,jingrui\}@illinois.edu} \\
}

\begin{document}

\renewcommand{\thefootnote}{\fnsymbol{footnote}}
\footnotetext[1]{Equal contribution.}
\renewcommand{\thefootnote}{\arabic{footnote}}

\maketitle

\addtocontents{toc}{\protect\setcounter{tocdepth}{0}} % such that the main part won't be in the content

\begin{abstract}
    Pretrained vision-language models such as CLIP achieve strong zero-shot generalization but remain vulnerable to distribution shifts caused by input corruptions. In this work, we investigate how corruptions affect CLIP’s image embeddings and uncover a consistent phenomenon we term as embedding variance collapse, where both intra-class and inter-class variances shrink as corruption severity increases. We find that this collapse is closely tied to performance degradation, with inter-class variance strongly correlated with classification accuracy. To explain this phenomenon, we analyze how corruptions alter the structure of the embedding space. Our theoretical results suggest that the visual encoder tends to encode corruption-related signals, which dilute class-discriminative features and compress the representation geometry. We further show that maximizing inter-class variance, even when estimated from pseudo-labels, can provably enhance embedding quality. Based on this insight, we propose {\algname}, a simple test-time adaptation method that maximizes pseudo-label-based inter-class variance on the fly using a mean accumulator and a gradient accumulator. {\algname} operates effectively with small batch sizes and consistently improves performance across multiple corruption benchmarks and CLIP architectures. Our code is available at \url{https://github.com/baowenxuan/Mint}. 
\end{abstract}
\section{Introduction}

Pretrained vision-language models (VLMs) such as CLIP \cite{clip} have demonstrated strong zero-shot generalization across a wide range of vision tasks \cite{regionclip,denseclip,styleclip}. However, their performance can degrade significantly under distribution shifts, such as common image corruptions \cite{corruption}. Test-time adaptation (TTA) has emerged as a promising strategy for improving model robustness under distribution shifts, by adapting the model during test-time without accessing source data or target labels \cite{tta_survey,tta_survey_2,tent}. This property makes TTA particularly suitable for the adaptation of pretrained VLMs, where the source training data is often large-scale, proprietary, or unavailable at deployment. 

Most existing TTA methods for VLMs focus on modifying the text prompt or embedding to improve image-text alignment \cite{tpt,difftpt,zpe,cupl,swapprompt,tps}, or leveraging similarities between different image embeddings to adapt the model prediction \cite{dmn,tda,dpe}. While these approaches achieve strong performance on standard benchmark datasets, they often overlook a key issue: the quality of image embeddings themselves can significantly degrade under corruption. Some recent methods \cite{clipartt,watt} attempt to address this by adjusting the image encoder’s normalization layers to align image-to-image or text-to-text similarities. However, such techniques typically require large batches to perform effective adaptation, making them unsuitable for many online TTA scenarios where only a few test samples are available at a time. Furthermore, these methods offer limited insight into why common corruptions cause accuracy degradation, and most lack theoretical analysis, making it difficult to understand when and why they succeed or fail. 

\begin{wrapfigure}{r}{0.5\textwidth}
  \centering
  \vspace{-2ex}
  \includegraphics[width=0.5\textwidth]{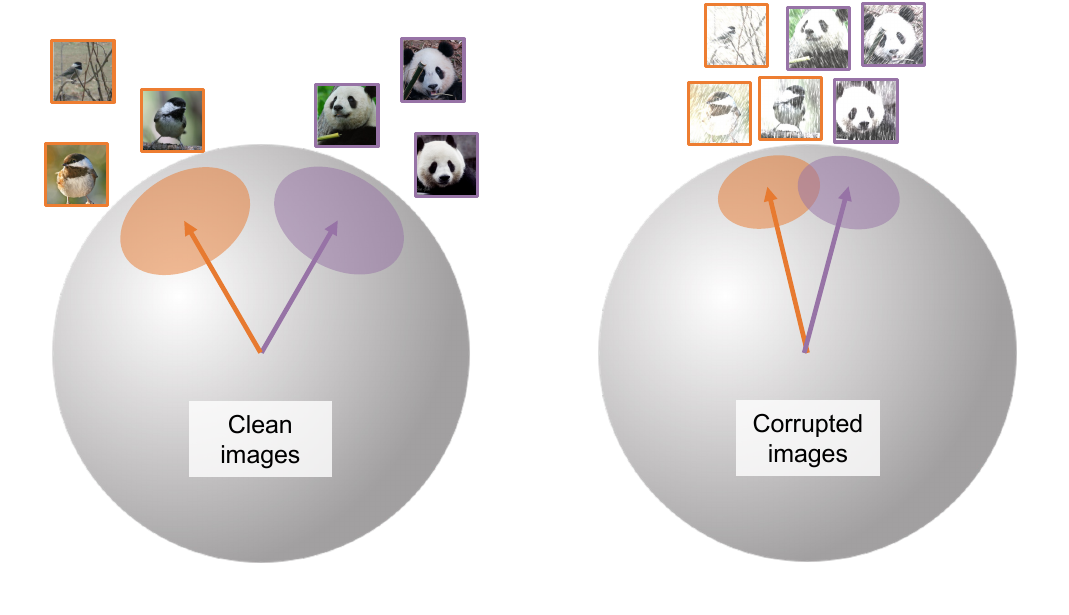}
  \vspace{-3ex}
  \caption{\textbf{Variance collapse.} Under corruptions, embeddings of images tend to become similar, regardless of whether they belong to the same class or different classes. }
  \label{fig:variance_collapse}
  \vspace{-1ex}
\end{wrapfigure}

In this work, we take a step back and ask: \textit{how exactly does corruption affect CLIP’s image embeddings?} To answer this, we evaluate the intra-class and inter-class variances of the embeddings using ground-truth labels, referred to as GT-intra and GT-inter, respectively. This analysis reveals a consistent and intriguing pattern: as corruption severity increases, both GT-intra and GT-inter variances consistently decrease. We term this phenomenon \textit{variance collapse}, which implies that under corruptions, embeddings of images tend to become more similar, regardless of whether they belong to the same class or different classes. The phenomenon is illustrated in Figure~\ref{fig:variance_collapse}. Moreover, we observe a strong correlation between inter-class variance and classification accuracy, suggesting that variance collapse is a key factor contributing to performance degradation.

To better understand and counteract variance collapse, we conduct a theoretical analysis of image embeddings' variances under distribution shifts. Our analysis shows that the simultaneous reduction of GT-intra and GT-inter variances can be attributed to the visual encoder projecting corruption-related patterns into the embedding space, which dilutes class-discriminative information. Our theoretical analysis further shows that even in the absence of ground-truth labels during adaptation, maximizing the inter-class variance computed from pseudo-labels (PL-inter) by updating the LayerNorm parameters can provably improve the quality of image embeddings and lead to more accurate classification. 

Motivated by this result, we design {\algname}, a simple test-time adaptation method that \textbf{m}aximizes the PL-\textbf{int}er variance on the fly. {\algname} is designed to operate reliably even when the batch size is extremely small, which is common in online adaptation settings. To enable stable adaptation under such constraints, {\algname} incorporates two key components: a mean accumulator and a gradient accumulator. The mean accumulator maintains cumulative averages of image embeddings for each pseudo-class and for the entire set of samples observed so far. This allows the estimation of PL-inter variance within each batch without requiring access to the full test set. In parallel, the gradient accumulator keeps track of the average update direction across batches, which reduces noise in parameter updates and improves adaptation stability. These two components together allow {\algname} to enhance class-discriminative signals and suppress corruption-related patterns in the image embedding space. We evaluate our method across a wide range of corruption benchmark datasets and CLIP architectures to demonstrate its robustness and generality. In all settings, {\algname} consistently outperforms existing TTA methods for VLMs, while also offering significant efficiency advantages. Our contributions are summarized as follows: 
\begin{itemize}
    \item We identify a phenomenon we refer to as variance collapse in CLIP image embeddings, where both intra-class and inter-class variances decrease as corruption severity increases.
    \item We provide a theoretical analysis that attributes this collapse to the visual encoder embedding corruption-related patterns, and show that maximizing PL-inter can improve embedding quality. 
    \item We propose {\algname}, a simple TTA method that maximizes PL-inter on the fly using a mean accumulator and a gradient accumulator, enabling effective adaptation even with extremely small batch sizes. 
    \item We demonstrate that {\algname} consistently improves the performance of CLIP models across multiple corruption benchmarks and architectures, outperforming existing TTA methods in both accuracy and efficiency. 
\end{itemize}
\section{Related works}

\textbf{Test-time adaptation (TTA)} adapts a source model to an unlabeled target domain during testing, without access to source data, making it suitable for pre-trained VLMs like CLIP. Early TTA methods for CLIP focus on modifying the text encoder or embedding, via prompt tuning \cite{tpt,difftpt,promptalign}, prompt weighting \cite{zpe}, or ensembling \cite{tip-adaptor,cupl}. Memory-based approaches \cite{tda,dmn,dpe,latte} store embeddings of high-confidence samples and use image similarity to guide predictions. Other training-free methods \cite{vte,zero} apply augmentations and confidence selection to enhance robustness without updating any model parameters. Although effective, most methods overlook a key issue: the quality of image embeddings degrades under corruptions. Recent approaches \cite{clipartt,watt} attempt to mitigate this by adjusting normalization layers, aligning the image-image and text-text similarities. However, they are very sensitive to batch size and introduce high computational overhead, limiting their use in online TTA settings.

\textbf{Inter-class separability} has been widely explored in supervised learning, where a common goal is to increase the distance between classes while reducing the variance within each class. A classic example is the Fisher score \cite{fisher}, which measures this separation and has been used for feature selection \cite{fisher_feat_select} and to improve domain adaptation by applying the Fisher criterion on the labeled source domains \cite{fisher_da}. However, computing such metrics typically requires access to ground-truth labels, which are unavailable in test-time adaptation. More recently, Matcha \cite{matcha} extended similar ideas to graph-based TTA by leveraging soft pseudo-labels. While effective, this method assumes simultaneous access to all nodes in the test graph, making it unsuitable for online adaptation scenarios where only small batches are available.

We provide a broader discussion of related works in Appendix \ref{appendix:discussion:related_works}. 
\section{Analysis} \label{sec:analysis}

\paragraph{Preliminary} CLIP \cite{clip} is a VLM consisting of an image encoder and a text encoder, which aligns images with their corresponding textual descriptions. Pretraining on a large-scale image-text dataset enables CLIP to perform zero-shot prediction. Specifically, for a classification task with $C$ classes, the text encoder embeds class descriptions (e.g., ``\texttt{a photo of a \{class\}}'') into normalized text embeddings $\mT = [\vt_1, \cdots, \vt_C]^\top \in \R^{C \times d}$, where $d$ is the embedding dimension. Given a test image, the image encoder produces a normalize image embedding $\vz_i \in \R^d$, and prediction is made via the maximum similarity score $\argmax_y \vz_i^\top \vt_y$. However, CLIP's performance degrades noticeably under common image corruptions \cite{clipartt,watt}, as its image encoder was not explicitly trained for robustness. 

%%%%%%%% %%%%%%%% %%%%%%%% %%%%%%%% %%%%%%%% %%%%%%%% %%%%%%%% %%%%%%%%

\subsection{Variance collapse}

In this subsection, we investigate how common corruptions affect the image embeddings extracted by CLIP's visual encoder, and how these changes influence classification accuracy. Motivated by Fisher score \cite{fisher,fisher_da,fisher_feat_select} and contrastive learning \cite{clip,infonce} objectives, we posit that high-quality image embeddings should exhibit low intra-class variance (i.e., samples from the same class are close) and high inter-class variance (i.e., samples from different classes are well separated). To formalize this intuition, given a target dataset with $C$ classes and $N$ images, we define the following variances: 
\begin{itemize}
    \item GT-total variance: $\gVGTtotal = \frac{1}{C} \sum_{c=1}^C \frac{\sum_{i=1}^N y_{ic}  \| \vz_i - \bar{\vz} \|_2^2}{\sum_{i=1}^N y_{ic}} $, 
    \item GT-inter variance: $\gVGTinter = \frac{1}{C} \sum_{c=1}^C \| \bar{\vz}_c - \bar{\vz} \|_2^2 $, 
    \item GT-intra variance: $\gVGTintra = \frac{1}{C} \sum_{c=1}^C \frac{\sum_{i=1}^N y_{ic}  \| \vz_i - \bar{\vz}_c \|_2^2}{\sum_{i=1}^N y_{ic}} $, 
\end{itemize}
where $\bar\vz = \frac{1}{N} \sum_{i=1}^N \vz_i$ is the average embedding for all images, $\bar\vz_c = \frac{\sum_{i=1}^N y_{ic} \vz_i}{\sum_{i=1}^N y_{ic}}$ is the average embedding of class $c$, and $\vy_i = [y_{i1}, \cdots, y_{iC}]^\top \in \{0, 1\}^C$ is the one-hot ground-truth label of image $i$, where $y_{ic} = 1$ if image $i$ corresponds to class $c$, and $y_{ic} = 0$ otherwise. Note that these variances are computed using the ground-truth labels. To distinguish them from the pseudo-label-based counterparts introduced later, we denote these metrics with a GT- prefix (e.g., GT-intra, GT-inter). 

\begin{figure}
    \centering
    \vspace{-2ex}
    \begin{minipage}[t]{0.48\linewidth}
        \centering
        \includegraphics[width=\linewidth]{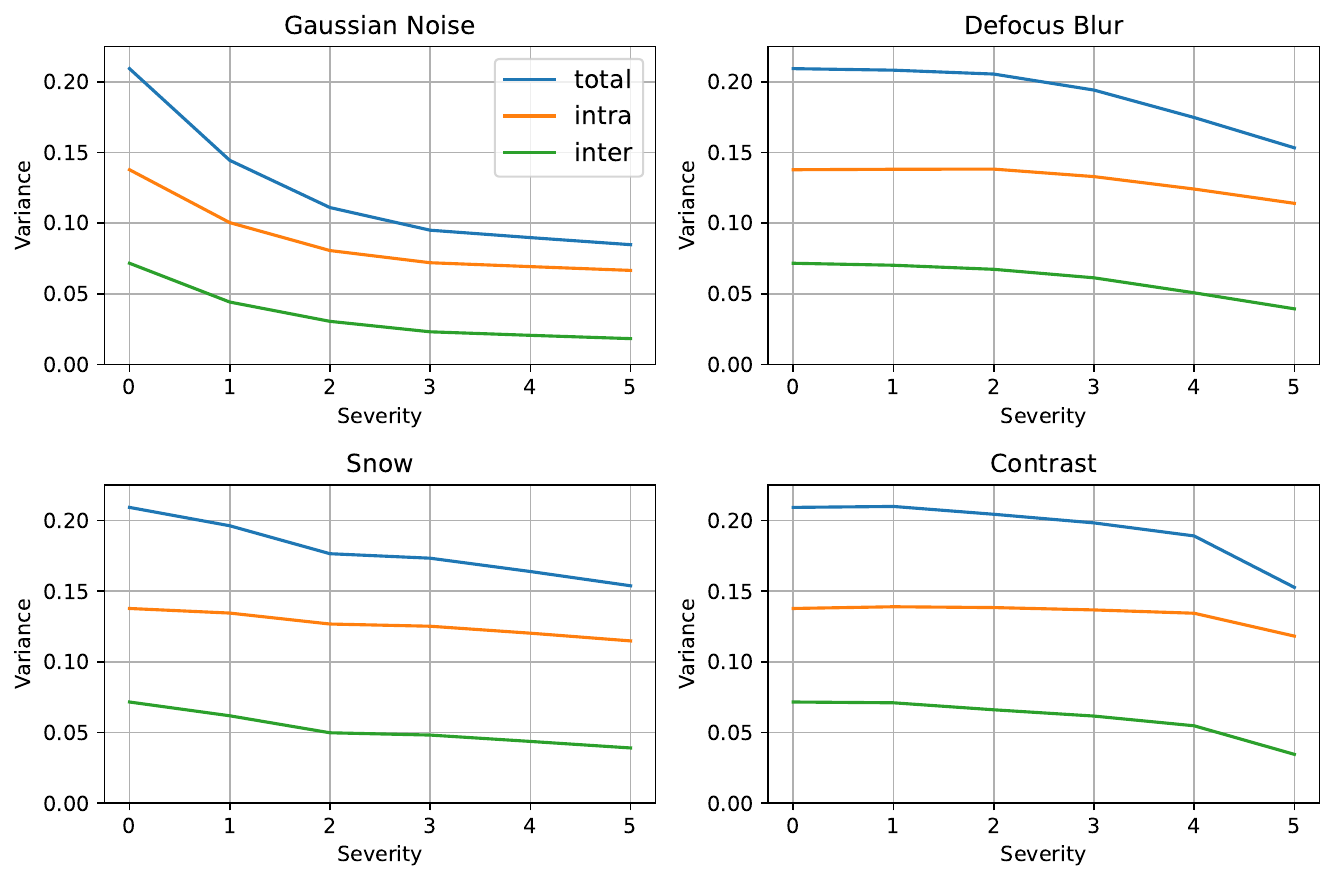}
        \vspace{-4ex}
        \caption{All types of variances decrease as the severity of corruption increases (severity=0 indicates original CIFAR-100 datasets without corruptions).}
        \label{fig:vardec_brief_cifar100}
    \end{minipage}%
    \hfill
    \begin{minipage}[t]{0.48\linewidth}
        \centering
        \includegraphics[width=\linewidth]{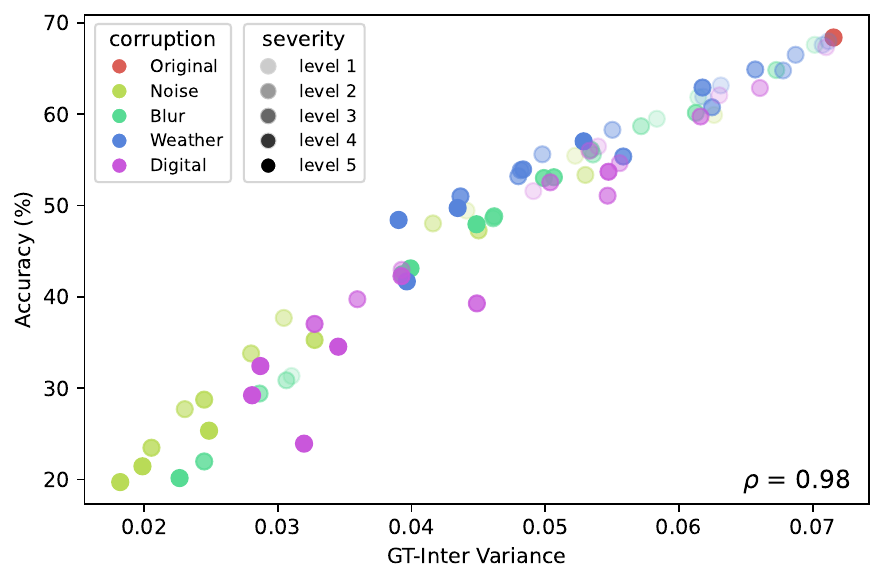}
        \vspace{-4ex}
        \caption{GT-inter variance is highly correlated with accuracy, with correlation 0.98. In comparison, the correlations between accuracy and GT-intra, GT-total are 0.86 and 0.94, respectively.}
        \label{fig:reg_brief_cifar100}
    \end{minipage}
    \vspace{-2ex}
\end{figure}

We compute the GT-total, GT-inter, and GT-intra variances on the corruption benchmark \cite{corruption}, which includes 15 types of common corruptions, each evaluated at 5 severity levels. Figure \ref{fig:vardec_brief_cifar100} presents the results on four representative corruptions. Additional results across all corruption types and datasets are included in Appendix \ref{appendix:corruption}. We observe that for all types of corruptions, both GT-intra and GT-inter variances consistently decrease as the severity increases. This indicates that the pairwise similarity between image embeddings increases under corruptions, regardless of whether the images belong to the same class or not. We refer to this phenomenon as \textbf{variance collapse}, where the image embeddings become increasingly indistinguishable under stronger corruptions. Furthermore, we find that variance collapse is closely linked to the drop in accuracy. We computed all three variance metrics along with the corresponding classification accuracy under 76 corruption settings (15 corruption types × 5 severity levels, plus the clean setting) on CIFAR-100-C. As shown in Figure~\ref{fig:reg_brief_cifar100}, the inter-class variance exhibits an extremely strong correlation with accuracy, indicating that this collapse could be a key factor driving the performance decline. 

%%%%%%%% %%%%%%%% %%%%%%%% %%%%%%%% %%%%%%%% %%%%%%%% %%%%%%%% %%%%%%%%

\subsection{Theoretical explanation}

In this subsection, we provide a theoretical explanation for the emergence of variance collapse. For clarity, we consider a balanced binary classification problem with $y_i \in \{0, 1\}$. Motivated by \cite{shift_analysis}, we assume that each image can be mapped to a disentangled latent representation $\vv_i = [\vvcls_i, \vvirr_i, \vvshift_i, \vvnoise_i] \in \R^d$, composed of four components: 
\begin{enumerate}
    \item \textit{Task-relevant features} $\vvcls_i$: 
        Semantic features that are directly predictive of the class label, $\vvcls_i = \vmu$ if $y_i = 1$ and $\vvcls_i = -\vmu$ if $y_i = 0$. 
    \item \textit{Task-irrelevant features} $\vvirr_i$: 
        Features unrelated to classification, such as background. It is preserved during pretraining due to CLIP's general representation learning objective. We assume $\vvirr_i \sim \text{Rademacher}^{\dirr}$, i.e., uniformly distributed in $\{-1, 1\}^{\dirr}$. 
    \item \textit{Structured distribution shift} $\vvshift_i$: 
        Features representing systematic distribution changes in the target domain, such as weather conditions or digital transforms. We assume $\vvshift_i = s \cdot \vdelta$, where $s$ indicates the severity of corruptions or distribution shifts. 
    \item \textit{Unstructured noise} $\vvnoise_i$: 
        Random noise introduced by the corruption process. We assume $\vvnoise_i \sim s \cdot \text{Rademacher}^{\dnoise}$, i.e., uniformly distributed in $\{-1, 1\}^{\dnoise}$. 
\end{enumerate}
Notice that by controlling the ratio of $s, \| \vmu \|_2, \| \vdelta \|_2$, we can freely adjust the ratio for each component. 
Following the structure of CLIP’s visual encoder, we assume that the latent representation $\vv$ first passes through a LayerNorm layer~\cite{layernorm} with a linear transformation, followed by normalization to unit length. For analytical simplicity, we omit the demeaning step in LayerNorm and ignore the bias term in its parameters,\footnote{This simplification is also known as RMSNorm~\cite{rmsnorm}.} under which the image embedding can be formulated as
\begin{align}
    \vz_i = \normalize \left( \frac{\vv_i}{\sqrt{\Var[\vv_i]}} \odot \vw \right), 
\end{align}
where $\odot$ represents element-wise multiplication of vectors, $\vw \in \R^{d}$ is the learnable weight of the LayerNorm layer, $\normalize(\cdot)$ denotes $\ell_2$ normalization. For simplicity, we assume $\vw = \vone$ at initialization. $\vw$ is updated during the adaptation. 

\begin{restatable}[Variance collapse]{theorem}{var}\label{thm:var}
    When the sample size $N \to +\infty$, 
    \begin{align}
        \gVGTinter \xrightarrow{p} \frac{\| \vmu \|_2^2}{\| \vmu \|_2^2 + \dirr + s^2 \cdot \| \vdelta \|_2^2 + s^2 \cdot \dnoise}, \ 
        \gVGTintra \xrightarrow{p} \frac{\dirr + s^2 \cdot \dnoise}{\| \vmu \|_2^2 + \dirr + s^2 \cdot \| \vdelta \|_2^2 + s^2 \cdot \dnoise}, 
    \end{align}
    where $s$ denotes the corruption severity. As $s$ increases, $\gVGTinter$ strictly decreases. In addition, $\gVGTintra$ also decreases when $\| \vdelta \|_2 \geq \sqrt{\dnoise / \dirr} \cdot \| \vmu \|_2$. 
\end{restatable}

Theorem \ref{thm:var} characterizes how the GT-inter and GT-intra variances change with increasing corruption severity. Combined with the empirical trends observed in Figure \ref{fig:vardec_brief_cifar100}, this suggests that common corruptions often induce significant structured distribution shifts, reflected as large $\vdelta$ in the latent space. As a result, the image encoder tends to embed corruption-related patterns into the representation itself. This dilutes class-discriminative features and introduces bias into the resulting image embeddings. 

\subsection{Maximization of inter variance}

Theorem \ref{thm:var} also supports that GT-inter variance has strong relevance to classification accuracy, as it reflects the proportion of task-relevant features within the overall feature representation. This insight motivates the idea that maximizing GT-inter variance should lead to improved classification accuracy under distribution shifts. However, several challenges arise in the context of TTA. First, the ground-truth labels are unavailable, so we must rely on pseudo-labels, i.e., the model's own prediction, which are noisy due to distribution shifts. Second, model updates in TTA are typically restricted to a small subset of parameters, such as LayerNorm weights, for better efficiency. In this part, we show that even under these constraints, using only pseudo-labels and updating only LayerNorm parameters, maximizing inter-class variance remains an effective and theoretically justified strategy for improving robustness to distribution shifts. 

\begin{restatable}[Maximization of PL-inter variance]{theorem}{adapt}\label{thm:adapt}
    When the sample size $N \to \infty$, 
    \begin{align}
        \gVPLinter \xrightarrow{p} \frac{C(\E \hat{y}_i)}{2} \cdot  \frac{4 \sigmayhy^2 \cdot \| \vmu \odot \vwcls \|_2^2 + \| \vsigmairr \odot \vwirr \|_2^2 +  \| \vsigmanoise \odot \vwnoise \|_2^2 }{\| \vmu \odot \vwcls \|_2^2 + \| \vwirr \|_2^2 + s^2 \cdot \| \vdelta \odot \vwshift \|_2^2 + s^2 \cdot \| \vwnoise \|_2^2}, 
    \end{align}
    where $C(\E \hat{y}_i) = \frac{1}{(\E \hat{y}_i)^2} + \frac{1}{(1 - \E \hat{y}_i)^2}$, $\sigmayhy = \Cov(y_i, \hat{y}_i)$, $\vsigmairr = \Cov(\vvirr, \hat{y}_i)$, and $\vsigmanoise = \Cov(\vvnoise, \hat{y}_i)$. Furthermore, when $\sigmayhy^2 \geq \frac{\| \vsigmairr \|_2^2}{4 \dirr}$ and $\sigmayhy^2 \geq \frac{\| \vsigmanoise \|_2^2}{4 \dnoise}$, we have
    \begin{align}
        \nabla_{\vwcls} \gVPLinter 
            &= C(\E \hat{y}_i) \cdot \frac{(4 \sigmayhy^2 \dirr - \| \vsigmairr \|_2^2) + 4 \sigmayhy^2 s^2 \| \vdelta \|_2^2 + (4 \sigmayhy^2 \dnoise - \| \vsigmanoise \|_2^2)}{( \| \vmu \|_2^2 + \dirr + s^2 \cdot (\| \vdelta \|_2^2 +  \dnoise) )^2} \cdot \vmu^2 \geq \vzero, \\
        \nabla_{\vwshift} \gVPLinter
            &= - C(\E \hat{y}_i) \cdot \frac{\gVPLinter}{\| \vmu \|_2^2 + \dirr + s^2 \cdot (\| \vdelta \|_2^2 +  \dnoise) } \cdot s^2 \cdot \vdelta^2 \leq \vzero. 
    \end{align}
\end{restatable}

This implies that when we perform a single gradient ascent step to maximize the PL-inter variance by updating the parameters of LayerNorm, the parameters associated with structured distribution shifts (i.e., $w_{\text{shift}}$) are necessarily suppressed. Furthermore, as long as the current prediction is reasonably accurate, meaning it depends more on task-relevant features than on task-irrelevant components or unstructured noise, maximizing PL-inter variance will increase the weights associated with task-relevant features (i.e., $w_{\text{cls}}$). As a result, this process reweighs the components in the final image embedding, enhancing the influence of task-relevant features while suppressing the effects of distribution shifts.

\section{Proposed method: \algname}

\begin{figure}
    \centering
    \vspace{-2ex}
    \includegraphics[width=1.0\linewidth]{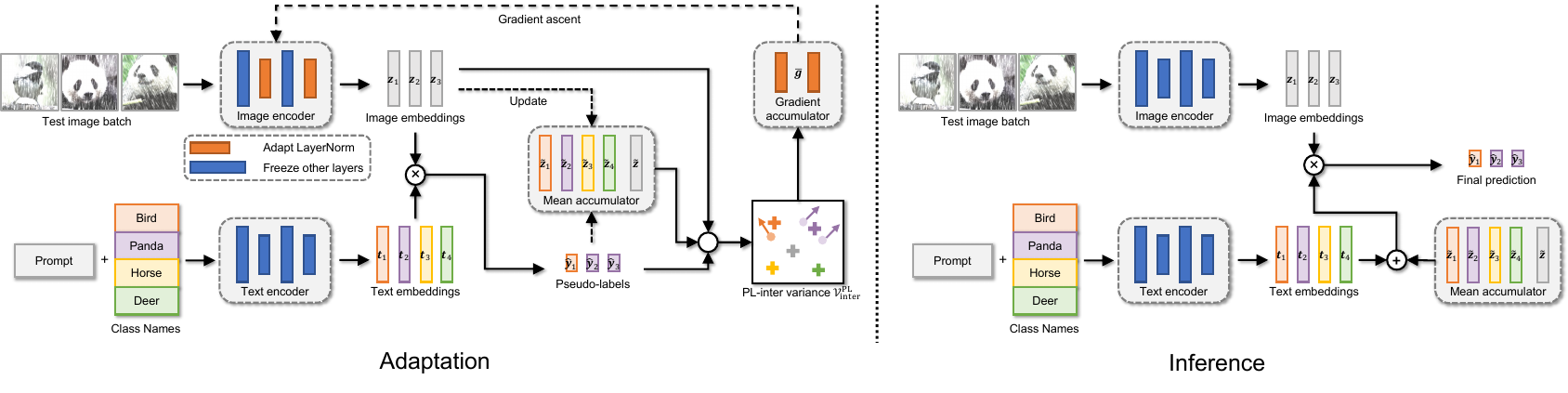}
    \vspace{-4ex}
    \caption{Overview of {\algname}. \textbf{Left: Adaptation phase.} Given a test image batch, we compute the PL-inter variance with the help of a mean accumulator and maximize it via gradient ascent. A gradient accumulator aggregates update directions across batches to robustly update the LayerNorm parameters. \textbf{Right: Inference phase.} The mean accumulator is used to adjust the text embeddings, and final predictions are made based on the similarity between image and text embeddings.}
    \label{fig:mint}
    \vspace{-2ex}
\end{figure}

In this section, we introduce our proposed algorithm {\algname}, which maximizes the PL-inter variance on the fly. While the previous section provides theoretical justification that maximizing PL-inter variance can improve test-time robustness, directly computing PL-inter variance requires access to the entire test dataset. However, in the online TTA setting, the model typically adapts using only a small batch or even a single sample at a time. This leads to noisy and potentially biased gradient directions that deviate from the true optimization target. To address this, we reparameterize the PL-inter variance and employ both a mean accumulator and a gradient accumulator to aggregate information across batches, enabling a more accurate approximation of the gradient in a streaming setting. Figure \ref{fig:mint} gives an overview of our method. 

%%%%%%%% %%%%%%%% %%%%%%%% %%%%%%%% %%%%%%%% %%%%%%%% %%%%%%%% %%%%%%%%

\subsection{Mean accumulator}

The most straightforward way to estimate PL-inter variance is to assume that the current batch fully represents the test data distribution and compute PL-inter using only samples in that batch. Under this approach, objectives across batches are computed independently. However, this naive strategy introduces significant noise, and even bias, into the PL-inter estimate. For instance, ImageNet-C contains 1,000 classes, but due to deployment and memory constraints, the test-time batch size is typically limited to just a few dozen samples. As a result, most classes present in a batch are represented by only a single sample, causing their estimated class means to degenerate into the samples themselves. In such cases, the distance between a sample and its class mean becomes zero, preventing us from estimating PL-intra variance. As a consequence, the computed objective ends up approximating PL-total variance rather than true PL-inter variance, which degrades adaptation performance under small batch sizes.

To address this issue, we first reparameterize PL-inter variance as the difference between PL-total and PL-intra variance. With detailed proof in Appendix \ref{appendix:analysis:var_decompose}, this decomposition can be written as:
\begin{align}
    \underbrace{\frac{1}{C}\sum_{c=1}^C \left\| \tilde{\vz}_c - \tilde{\vz} \right\|_2^2}_{\gVPLinter} 
    = \underbrace{\frac{1}{C}\sum_{c=1}^C \frac{\sum_{i=1}^N \hat{y}_{ic} \left\| \vz_i - \tilde{\vz} \right\|_2^2}{\sum_{i=1}^N \hat{y}_{ic}}}_{\gVPLtotal} 
    - \underbrace{\frac{1}{C}\sum_{c=1}^C \frac{\sum_{i=1}^N \hat{y}_{ic} \left\| \vz_i - \tilde{\vz}_c \right\|_2^2}{\sum_{i=1}^N \hat{y}_{ic}}}_{\gVPLintra}, 
    \label{eq:PL_decompose}
\end{align}
where $\vz_i$ is the embedding for $i$-th image, $\tilde{\vz} = \frac{1}{N}\sum_{i=1}^N \vz_i$ is the global average embedding, and $\tilde{\vz}_c = \frac{\sum_{i=1}^N \hat{y}_{ic} \vz_i}{\sum_{i=1}^N \hat{y}_{ic}}$ is the average embedding of all images predicted as class $c$ by CLIP. This reformulation reveals that maximizing PL-inter is equivalent to jointly maximizing PL-total variance and minimizing PL-intra variance, encouraging each embedding $\vz_i$ to move away from the global mean $\tilde{\vz}$ and toward its corresponding class mean $\tilde{\vz}_c$, with the gradient direction approximately given by $\tilde{\vz}_c - \tilde{\vz}$ when the sample size is sufficiently large. This insight suggests that more accurate estimates of $\tilde{\vz}$ and $\tilde{\vz}_c$ can lead to better gradient directions. Therefore, instead of estimating these means using only the current batch, we use a mean accumulator to maintain cumulative averages $\tilde{\vz}$ and $\{\tilde{\vz}_c\}_{c=1}^C$. Every time when we observe a new image with embedding $\vz_i$ and CLIP's prediction $\hat{y}_i$ as pseudo-label, 
\begin{align}
    \tilde{\vz} \leftarrow \frac{K}{K + 1} \tilde{\vz} + \frac{1}{K + 1} \vz_i, \quad
    \tilde{\vz}_{\hat{y}_i} \leftarrow \frac{K_{\hat{y}_i}}{K_{\hat{y}_i} + 1} \tilde{\vz}_{\hat{y}_i} + \frac{1}{K_{\hat{y}_i} + 1} \vz_i, 
\end{align}
where $K$ is the total number of seen samples, and $K_{\hat{y}_i}$ is the number of seen samples with pseudo-label $\hat{y}_i$. After replacing the class and global means in Equation (\ref{eq:PL_decompose}) with the cumulative averages, the final objective for the $b$-th batch $\sB_b$ becomes
\begin{align}
    \gVPLinter(\sB_b) = \frac{1}{C_b} \sum_{c=1}^{C_b} \frac{\sum_{i \in \sB_b} \hat{y}_{ic} \left\| \vz_i - \tilde{\vz} \right\|_2^2}{\sum_{i \in \sB_b} \hat{y}_{ic}} - \frac{1}{C_b} \sum_{c=1}^{C_b}\frac{\sum_{i \in \sB_b} \hat{y}_{ic} \left\| \vz_i - \tilde{\vz}_c \right\|_2^2}{\sum_{i \in \sB_b} \hat{y}_{ic}}, 
\end{align}
where $C_b$ is the number of unique classes present in batch $\sB_b$.

%%%%%%%% %%%%%%%% %%%%%%%% %%%%%%%% %%%%%%%% %%%%%%%% %%%%%%%% %%%%%%%%

\subsection{Gradient accumulator}

While the mean accumulator mitigates systematic bias in the objective by stabilizing the estimates of class and global means, it does not eliminate the noise in the individual gradient contributions from $\vz_i$, which are still computed over the current batch. 
To further reduce gradient estimation error, we introduce a simple gradient accumulator that mimics adaptation with a larger effective batch size. Specifically, for the $b$-th batch, if the gradient computed on the current batch is $\vg_b$, we maintain a cumulative average of gradients $\bar{\vg}$ over the seen $b$ batches:
\begin{align}
    \bar{\vg} \leftarrow \frac{b - 1}{b} \bar{\vg} + \frac{1}{b} \vg_b, 
\end{align}
and update the LayerNorm parameters in the direction of $\bar{\vg}$. We perform only a single step of update on each batch. 

%%%%%%%% %%%%%%%% %%%%%%%% %%%%%%%% %%%%%%%% %%%%%%%% %%%%%%%% %%%%%%%%

\subsection{Adjust text embedding}

In addition to estimating PL-inter variance, the cumulative class means can also be leveraged to adjust the text embeddings, thereby improving alignment between the image and text modalities. Motivated by prior works in TTA \cite{t3a} and Bayesian estimation, we adopt a simple, training-free approach to refine the text embeddings using accumulative embedding means. 
Specifically, we maintain a separate mean accumulator to store the image embeddings produced by the adapted image encoder. The refinement of text embeddings is given by
\begin{align}
    \tilde{\vt}_c \leftarrow \normalize \left(\frac{\Kprior}{\Kprior + K} \cdot \vt_c + \frac{K}{\Kprior + K} \cdot \tilde{\vz}_c \right), \quad c = 1, \cdots, C, 
\end{align}
where $\Kprior$ is a hyperparameter controlling the strength of prior. This design enables dynamic adjustment of the text embedding. In the early stage of adaptation, the image embedding means may be less reliable, so we assign more weight to the original text embedding $\vt_c$. As adaptation progresses and the quality of the estimated class-wise means $\tilde{\vz}_c$ improves, we gradually place more weight on $\tilde{\vz}_c$. % \bao{Practically we use another mean accumulator for this part. However this may introduce unnecessary confusion, so I plan to omit this detail. }

The final prediction is given by $\argmax_y \vz_i^\top \tilde{\vt}_y$. After making prediction on each batch, we reset both the image encoder and the optimizer state to their initial values. However, the mean accumulator and gradient accumulator are preserved and carried over to the next batch, allowing information aggregated from previous samples to guide the adaptation on subsequent inputs.

\section{Experiments} \label{sec:experiment}

In this section, we use experiments to answer the following research questions
\begin{itemize}
    \item \textbf{RQ1}: Can {\algname} effectively improve the performance of CLIP models under common corruptions, especially in low batch size scenarios? 
    \item \textbf{RQ2}: Does {\algname} effectively mitigate the variance collapse?
    \item \textbf{RQ3}: How efficient is {\algname} in terms of computational time? 
\end{itemize}

\paragraph{Setup and baselines}
We test {\algname} with different combination of model architectures and corruption datasets \cite{corruption}: ViT-B/32 \cite{vit} on CIFAR-10-C \cite{cifar}, ViT-B/16 on CIFAR-100-C, and ViT-L/14 on ImageNet-C \cite{imagenet}, all with corruption severity of 5. We consider a standard TTA setting, where the model is adapted to each type of corruption independently. We compare {\algname} with a wide range of existing TTA methods designed for VLMs. VTE \cite{vte} and Zero \cite{zero} aggregate image embeddings from multiple augmentations. TPT \cite{tpt} and TPS \cite{tps} minimize the marginal entropy to encourage consistency across augmented views. TDA \cite{tda} and DMN-ZS \cite{dmn} leverage sample-wise similarity to adjust predictions. WATT-S \cite{watt} and CLIPArTT \cite{clipartt} improve modality alignment by aligning image-to-image and text-to-text similarities. Unless otherwise specified, we use a default batch size of 20 during adaptation. {\algname} uses Adam \cite{adam} optimizer with learning rate $0.007$ for ViT-B models and $0.015$ for ViT-L/14, and $\Kprior = 10{,}000$. Hyperparameter settings for baselines are provided in the Appendix \ref{appendix:exp:detail}. 

%%%%%%%% %%%%%%%% %%%%%%%% %%%%%%%% %%%%%%%% %%%%%%%% %%%%%%%% %%%%%%%%

\begin{table}
    \centering
    \caption{Mean accuracy (\%) on corruption benchmarks. Error bars are deferred to Appendix \ref{appendix:exp:rq1}. }
    \label{tab:acc:main}
    % \vspace{1ex}
    \resizebox{1.0\linewidth}{!}{
    \setlength{\tabcolsep}{1.0mm}{
        \begin{tabular}{llcccccccccccccccc}
            \toprule
            & & \multicolumn{16}{c}{ViT-B/32 on CIFAR-10-C} \\
            \cmidrule(lr){3-18}
            Method & Venue & \multicolumn{3}{c}{Noise} & \multicolumn{4}{c}{Blur} & \multicolumn{4}{c}{Weather} & \multicolumn{4}{c}{Digital} & \multirow{2.5}{*}{Avg.} \\
            \cmidrule(lr){3-5} \cmidrule(lr){6-9} \cmidrule(lr){10-13} \cmidrule(lr){14-17}
            & & Gauss. & Shot & Impul. 
            & Defoc. & Glass & Motion & Zoom  
            & Snow & Frost & Fog & Brit. 
            & Contr. & Elastic & Pixel & JPEG \\
            \midrule
            CLIP \cite{clip} & ICML'21 & 
                35.5 & 40.0 & 43.2 & 70.0 & 41.4 & 64.5 & 70.2 & 70.8 & 72.3 & 66.7 & 81.4 & 64.5 & 59.6 & 48.2 & 56.7 & 59.0 \\
            Ensemble & - & 
                38.8 & 42.7 & 42.8 & 72.6 & 43.9 & 66.8 & 71.7 & 73.9 & 75.8 & 68.9 & 83.7 & 67.2 & 61.9 & 51.8 & 58.6 & 61.4 \\
            % Tent \cite{tent} & ICLR'21 & 24.4 & 29.0 & 34.7 & 79.5 & 38.8 & 70.2 & 78.4 & 81.3 & 81.7 & 79.4 & 88.6 & 81.5 & 72.2 & 61.4 & 54.6 & 63.7\\
            TPT \cite{tpt} & NeurIPS'22 & 
                42.9 & 46.2 & 47.1 & 71.5 & 46.4 & 68.1 & 72.7 & 73.7 & 75.9 & 68.9 & 83.7 & 73.9 & 62.5 & 50.3 & 58.2 & 62.8 \\
            TDA \cite{tda} & CVPR'24 & 41.2 & 44.1 & 43.3 &	73.9 & 45.1 & 68.1 & 73.6 & 74.0 & 76.7 & 69.6 & 84.0 & 66.6 & 62.3 & 54.7 & 58.4 & 62.4\\
            DMN-ZS \cite{dmn} & CVPR'24 & 37.6 & 41.5 & 42.5 & 69.4 & 43.8 & 65.9 & 70.5 & 70.2 & 71.2 & 64.0 & 80.7 & 58.6 & 59.4 & 54.9 & 58.1 & 59.2\\
            VTE \cite{vte} & ECCV-W’24 & 47.6 & 50.5	& 49.8 & 70.4 & 49.8	& 70.2 & 73.4 & 74.4 & 77.3	& 71.4 & 83.6 & \textbf{81.2} & 65.5 & 55.3 & 58.8 & 65.3\\
            Zero \cite{zero} & NeurIPS'24 & 47.9 & 50.5 & 50.0 & 70.3 & 50.3	& 69.7 & 73.6 & 74.5 & 77.1 & 71.5 & 83.5 & 80.6 & 66.0 & 55.2 & 58.9 & 65.3\\
            WATT-S \cite{watt} & NeurIPS'24 & 
                53.2 & 54.9 & 50.7 & 75.0 & 55.4 & 71.1 & 74.8 & 75.4 & 77.0 & 72.7 & 84.2 & 73.1 & 65.4 & 61.1 & 62.3 & 67.1 \\
            TPS \cite{tps} & WACV'25 & 
                45.5 & 49.4 & 49.2 & 73.8 & 50.7 & 71.4 & 76.0 & 77.0 & \textbf{79.2} & 73.3 & 85.3 & 79.5 & 67.2 & 56.6 & 61.8 & 66.4 \\
            CLIPArTT \cite{clipartt} & WACV'25 & 
                45.2 & 48.7 & 47.1 & 73.4 & 49.9 & 69.0 & 73.0 & 74.1 & 76.2 & 70.1 & 84.3 & 71.4 & 64.1 & 58.5 & 60.5 & 64.4 \\
            {\algname} & - & \textbf{59.0} & \textbf{62.4} & \textbf{54.2} & \textbf{75.8} & \textbf{61.8} & \textbf{77.1} & \textbf{78.9} & \textbf{79.0} & 78.9 & \textbf{75.2} & \textbf{86.3} & 76.9 & \textbf{70.1} & \textbf{66.6} & \textbf{63.4} & \textbf{71.0} \\

            %%%%%%%% %%%%%%%% %%%%%%%% %%%%%%%% %%%%%%%% %%%%%%%% %%%%%%%% %%%%%%%% 
            \midrule 
            %%%%%%%% %%%%%%%% %%%%%%%% %%%%%%%% %%%%%%%% %%%%%%%% %%%%%%%% %%%%%%%% 

            & & \multicolumn{16}{c}{ViT-B/16 on CIFAR-100-C} \\
            \cmidrule(lr){3-18}
            Method & Venue & \multicolumn{3}{c}{Noise} & \multicolumn{4}{c}{Blur} & \multicolumn{4}{c}{Weather} & \multicolumn{4}{c}{Digital} & \multirow{2.5}{*}{Avg.}  \\
            \cmidrule(lr){3-5} \cmidrule(lr){6-9} \cmidrule(lr){10-13} \cmidrule(lr){14-17}
            & & Gauss. & Shot & Impul. 
            & Defoc. & Glass & Motion & Zoom  
            & Snow & Frost & Fog & Brit. 
            & Contr. & Elastic & Pixel & JPEG \\
            \midrule
            CLIP \cite{clip} & ICML'21 & 
                19.7 & 21.4 & 25.3 & 42.5 & 20.2 & 43.1 & 48.0 & 48.4 & 49.7 & 41.7 & 57.0 & 34.5 & 29.2 & 23.9 & 32.4 & 35.8 \\
            Ensemble & - & 
                22.9 & 24.3 & 29.6 & 43.6 & 20.1 & 43.7 & 48.7 & 48.9 & 50.3 & 41.8 & 58.1 & 35.2 & 29.2 & 26.3 & 33.6 & 37.1 \\
            % Tent \cite{tent} & ICLR'21 & 10.6 & 14.1 & 21.1 & 51.8 & 7.8 & 52.0 & 55.4 & 55.1 & 48.3 & 50.8 & 65.8 & 53.0 & 33.0 & 42.9 & 39.6 & 40.1\\
            TPT \cite{tpt} & NeurIPS'22 & 
                17.3 & 19.2 & 25.6 & 42.4 & 20.0 & 42.2 & 47.9 & 49.0 & 50.0 & 42.7 & 57.5 & 38.0 & 30.3 & 25.5 & 32.5 & 36.0 \\
            TDA \cite{tda} & CVPR'24 & 23.8	& 26.0 & 32.5 & 45.7 & 21.5 & 44.4	& 50.5 & 49.6 & 51.5 & 42.8 & 59.2 & 36.8 & 29.7 & 28.1 & 34.3 & 38.4\\
            DMN-ZS \cite{dmn} & CVPR'24 & 23.9 & 25.6 & 31.7 & 45.5 & 21.6 & 45.0 & 51.1 & 49.6 & 52.0 & 43.0 & 60.3 & 36.0 & 30.5 & 27.5 & 34.7 & 38.5\\
            VTE \cite{vte} & ECCV-W’24 & 20.2 & 21.2 & 28.4 & 39.9 & 18.5 & 39.0 & 44.7 & 47.6 & 48.8 & 43.2 & 55.7 & 49.9 & 30.4 & 30.3 & 30.6 & 36.6\\
            Zero \cite{zero} & NeurIPS'24 & 19.9 & 21.5 & 29.6 & 40.4 & 18.5 & 39.6 & 44.8 & 47.8 & 48.3 & 43.3 & 55.8 & \textbf{50.0} & 30.6 & 30.4 & 30.7 & 36.8\\
            WATT-S \cite{watt} & NeurIPS'24 & 
                27.5 & 29.8 & 36.4 & 47.5 & 26.8 & 46.8 & 51.6 & 51.6 & 52.3 & 46.6 & 61.0 & 43.5 & 34.3 & \textbf{35.9} & 37.3 & 41.9 \\
            TPS \cite{tps} & WACV'25 & 
                22.6 & 24.4 & 31.0 & 44.0 & 20.1 & 43.6 & 49.0 & 50.5 & 51.3 & 44.3 & 59.1 & 45.1 & 30.6 & 28.8 & 33.8 & 38.6 \\
            CLIPArTT \cite{clipartt} & WACV'25 & 
                24.9 & 27.1 & 32.5 & 47.4 & 23.4 & 47.2 & 52.0 & 51.6 & \textbf{52.5} & 46.5 & 61.2 & 41.2 & 33.7 & 32.6 & 37.0 & 40.7 \\
            {\algname} & - & \textbf{29.4} & \textbf{30.8} & \textbf{38.6} & \textbf{50.7} & \textbf{27.1} & \textbf{49.9} & \textbf{55.5} & \textbf{53.0} & 51.8 & \textbf{50.6} & \textbf{65.6} & 48.1 & \textbf{36.8} & 34.4 & \textbf{38.7} & \textbf{44.1} \\

            %%%%%%%% %%%%%%%% %%%%%%%% %%%%%%%% %%%%%%%% %%%%%%%% %%%%%%%% %%%%%%%% 
            \midrule 
            %%%%%%%% %%%%%%%% %%%%%%%% %%%%%%%% %%%%%%%% %%%%%%%% %%%%%%%% %%%%%%%% 

            & & \multicolumn{16}{c}{ViT-L/14 on ImageNet-C} \\
            \cmidrule(lr){3-18}
            Method & Venue & \multicolumn{3}{c}{Noise} & \multicolumn{4}{c}{Blur} & \multicolumn{4}{c}{Weather} & \multicolumn{4}{c}{Digital} & \multirow{2.5}{*}{Avg.} \\
            \cmidrule(lr){3-5} \cmidrule(lr){6-9} \cmidrule(lr){10-13} \cmidrule(lr){14-17}
            & & Gauss. & Shot & Impul. 
            & Defoc. & Glass & Motion & Zoom  
            & Snow & Frost & Fog & Brit. 
            & Contr. & Elastic & Pixel & JPEG \\
            \midrule
            CLIP \cite{clip} & ICML'21 & 
                27.4 & 29.4 & 28.7 & 34.6 & 25.3 & 41.0 & 36.7 & 49.8 & 44.1 & 49.7 & 65.4 & 35.1 & 30.3 & 53.5 & 42.2 & 39.6 \\
            Ensemble & - & 
                29.1 & 30.4 & 30.1 & 37.5 & 27.2 & 44.2 & 39.2 & 52.4 & 46.4 & 52.7 & 67.8 & 34.5 & 32.4 & 56.2 & 44.3 & 41.6 \\
            % Tent \cite{tent} & ICLR'21 & 34.3 & 35.7 & 35.6 & 40.2 & 34.7 & 46.9 & 42.5 & 55.1 & 47.1	& 55.1 & 68.5 & 43.0 & 38.4 & 57.8 & 49.2 & 45.6\\
            TPT \cite{tpt} & NeurIPS'22 & 
                27.2 & 29.1 & 29.3 & 35.7 & 26.6 & 41.1 & 38.1 & 51.4 & 46.3 & 51.6 & 67.7 & 39.4 & 32.1 & 55.9 & 45.5 & 41.1 \\
            TDA \cite{tda} & CVPR'24 &  29.1 & 30.5 & 31.0 & 37.7 & 28.0 & 44.5 & 39.5 & 53.4 & 47.8 & 53.6 & 68.3 & 36.8 & 33.3 & 56.7 & 44.4 & 42.3\\
            DMN-ZS \cite{dmn} & CVPR'24 & 29.0 & 30.4 & 30.4 & 37.5 & 27.3 & 44.3 & 39.3 & 52.5 & 46.7 & 52.7 & 67.8 & 34.9 & 32.4 & 56.2 & 44.3 & 41.7\\
            VTE \cite{vte} & ECCV-W’24 & 23.2 & 26.5 & 24.9 & 34.5 & 25.8 & 39.7 & 38.2 & 49.0 & 45.7 & 49.8 & 67.0 & 44.4 & 32.1 & 55.8 & 46.5 & 40.2\\
            Zero \cite{zero} & NeurIPS'24 & 24.1 & 26.9 & 25.8 & 35.8 & 26.9 & 40.3 & 39.4 & 49.5	& 46.2 & 50.7 & 66.8 & 44.9	& 32.6 & 56.4 & 47.4 & 40.9\\
            WATT-S \cite{watt} & NeurIPS'24 & 
                31.7 & 33.5 & 34.6 & 38.7 & 31.3 & 45.2 & 41.2 & 52.7 & 47.8 & 54.5 & 67.5 & 42.9 & 34.8 & 56.3 & 45.9 & 43.9 \\
            TPS \cite{tps} & WACV'25 & 
                28.9 & 31.0 & 30.7 & 37.8 & 28.0 & 43.4 & 40.8 & 53.3 & \textbf{47.9} & 53.5 & \textbf{69.2} & 43.8 & 33.3 & 57.3 & 47.0 & 43.1 \\
            CLIPArTT \cite{clipartt} & WACV'25 & 
                29.2 & 31.0 & 30.8 & 34.5 & 28.1 & 41.9 & 38.0 & 49.9 & 44.7 & 50.1 & 64.5 & 39.2 & 32.4 & 53.0 & 42.4 & 40.7 \\
            {\algname} & - & \textbf{33.0} & \textbf{34.3} & \textbf{37.3} & \textbf{39.6} & \textbf{37.2} & \textbf{46.6} & \textbf{45.1} & \textbf{55.2} & 46.6 & \textbf{57.5} & 67.7	& \textbf{48.9} & \textbf{43.9} & \textbf{58.2} & \textbf{54.6} & \textbf{47.0}\\
            \bottomrule 
        \end{tabular}
    }
    }
\end{table}

\paragraph{Main results (RQ1)}
The experimental results are summarized in Table \ref{tab:acc:main}. We observe that training-free methods generally perform worse, as they do not update the model during adaptation. Among them, TPS achieves relatively strong performance by adjusting the text embeddings. CLIPArTT and WATT-S, which allow updates to the image encoder, perform best among the baselines. However, these methods do not share information across batches, which limits their overall effectiveness. Across all settings, {\algname} consistently improves accuracy and achieves the best performance. Compared to the strongest baselines, {\algname} yields absolute gains of 3.9\%, 2.2\%, and 3.1\%, respectively. 

\begin{table}
    \centering
    \caption{Accuracy (\meansd{mean}{s.d.} \%) of {\algname} with various batch size.}
    \label{tab:batch_brief}
    \resizebox{1.0\linewidth}{!}{
    \begin{tabular}{llccccccccc}
        \toprule
        \multirow{2.5}{*}{Architecture} & \multirow{2.5}{*}{Dataset} & \multirow{2.5}{*}{CLIP} & \multicolumn{8}{c}{\algname} \\
        \cmidrule(lr){4-11}
        & & & $\text{BS}=1$ & $\text{BS}=2$ & $\text{BS}=5$ & $\text{BS}=10$ & $\text{BS}=20$ & $\text{BS}=50$ & $\text{BS}=100$ & $\text{BS}=200$ \\
        \midrule
        ViT-B/32 & CIFAR-10-C  & 59.0 & \meansd{70.5}{0.1} & \meansd{70.5}{0.1} & \meansd{71.0}{0.0} & \meansd{71.0}{0.1} & \meansd{71.0}{0.1} & \meansd{71.0}{0.1} & \meansd{70.9}{0.1} & \meansd{70.6}{0.1} \\
        ViT-B/16 & CIFAR-100-C & 35.8 & \meansd{43.1}{0.1} & \meansd{43.1}{0.1} & \meansd{43.3}{0.1} & \meansd{43.6}{0.1} & \meansd{44.1}{0.1} & \meansd{44.5}{0.1} & \meansd{44.5}{0.1} & \meansd{44.6}{0.1} \\
        ViT-L/14 & ImageNet-C  & 39.6 & \meansd{45.8}{0.1} & \meansd{46.2}{0.1} & \meansd{46.7}{0.1} & \meansd{46.8}{0.1} & \meansd{47.0}{0.2} & \meansd{47.1}{0.1} & \meansd{47.0}{0.2} & \meansd{46.8}{0.1}  \\
        \bottomrule
    \end{tabular}
    }% }
\end{table}

\paragraph{Robustness to batch size (RQ1)}
To evaluate the robustness of {\algname} under different test-time conditions, we run it with batch sizes ranging from 1 to 200, using the same set of hyperparameters across all settings. As shown in Table \ref{tab:batch_brief}, {\algname} consistently maintains strong performance across this range. Even in the extreme case of batch size 1, it achieves significant accuracy gains, demonstrating its effectiveness in highly constrained online adaptation scenarios.

%%%%%%%% %%%%%%%% %%%%%%%% %%%%%%%% %%%%%%%% %%%%%%%% %%%%%%%% %%%%%%%%

\begin{figure}
    \centering
    \vspace{-2ex}
    \includegraphics[width=1.0\linewidth]{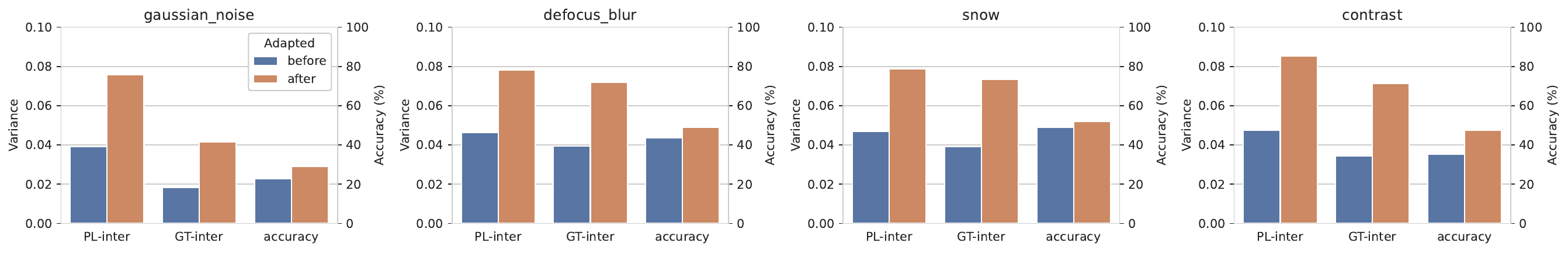}
    \vspace{-2ex}
    \caption{{\algname} alleviates variance collapse. }
    \label{fig:changes_brief}
    % \vspace{-2ex}
\end{figure}

\paragraph{Variance collapse (RQ2)}
We investigate the underlying mechanism of {\algname} by analyzing its effect on the image embeddings. Specifically, we evaluate the PL-inter variance, GT-inter variance, and classification accuracy on CIFAR-100-C before and after adaptation, under four representative types of corruption (same as Figure \ref{fig:vardec_brief_cifar100}). As shown in Figure \ref{fig:changes_brief}, {\algname} successfully increases PL-inter variance by design, and this also leads to a clear improvement in GT-inter variance. The increased GT-inter variance is accompanied by a rise in accuracy, indicating that {\algname} effectively mitigates variance collapse. Additional results across all corruption types and datasets are provided in Appendix~\ref{appendix:exp:rq2}. 

%%%%%%%% %%%%%%%% %%%%%%%% %%%%%%%% %%%%%%%% %%%%%%%% %%%%%%%% %%%%%%%%

\begin{wraptable}{r}{0.45\textwidth}
    \centering
    \vspace{-4ex}
    \caption{Comparison of testing time. } 
    \label{tab:time}
    \vspace{1ex}
    \scriptsize{
    \setlength{\tabcolsep}{1.0mm}{
    \begin{tabular}{lccc}
        \toprule
        Method   & Testing Time & Accuracy (\%) & Gain (\%) \\
        \midrule
        CLIP      & 21s    & 35.8 & --   \\
        TPT       & 23m21s & 36.0 & +0.2 \\
        VTE       & 9m45s  & 36.6 & +0.8 \\
        Zero      & 9m50s  & 36.8 & +1.0 \\
        TDA       & 33s    & 38.4 & +2.6 \\
        DMN-ZS    & 30s    & 38.5 & +2.7 \\
        TPS       & 9m58s  & 38.6 & +2.8 \\
        CLIPArTT  & 7m40s  & 40.7 & +4.9 \\
        WATT\mbox{-}S & 50m20s & 41.9 & +6.1 \\
        {\algname}      & 1m07s  & \textbf{44.1} & \textbf{+8.3} \\
        \bottomrule
    \end{tabular}
    }
    }
    \vspace{-6ex}
\end{wraptable}

% WATT-S: 
% CLIPARTT: 

    % \begin{tabular}{lrcc}
    %     \toprule
    %     Method & Testing Time & Accuracy (\%) & Gain (\%) \\
    %     \midrule
    %     CLIP            & 21s       & 35.8  & - \\
    %     TPT             & 23m 21s   & 36.0  & +0.2 \\
    %     TPS             & 9m 58s    & 38.6  & +2.8 \\
    %     % Tent            & 41s       & 40.1  & +4.3 \\
    %     CLIPArTT        & 7m 40s    & 40.7  & +4.9 \\
    %     WATT-S          & 50m 20s   & 41.9  & +6.1 \\
    %     \algname        & 1m 07s    & 44.1  & +8.3 \\
    %     % \algname-online & 48s\\
    %     \bottomrule
    % \end{tabular}
\paragraph{Efficiency (RQ3)}
We compare the testing time of {\algname} with baseline algorithms on CIFAR-100-C by measuring the time required to process one corruption type (10,000 images). As shown in Table \ref{tab:time}, Mint runs substantially faster than other training-based TTA methods. This efficiency primarily stems from its simple design and the fact that it performs only a single model update per batch, unlike methods that require multiple iterative updates during adaptation. Notably, {\algname} is only slower than CLIP and other training-free and augmentation-free baselines.

%%%%%%%% %%%%%%%% %%%%%%%% %%%%%%%% %%%%%%%% %%%%%%%% %%%%%%%% %%%%%%%%

\begin{figure}
    \centering
    % \vspace{-2ex}
    \hspace{0.02\linewidth}
    \begin{minipage}[t]{0.38\linewidth}
        \centering
        \includegraphics[width=\linewidth]{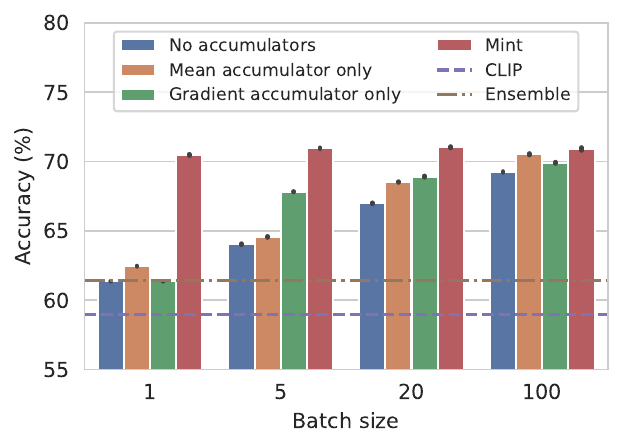}
        \vspace{-4ex}
        \caption{Ablation study. }
        \label{fig:ablation}
    \end{minipage}%
    \hfill
    \begin{minipage}[t]{0.53\linewidth}
        \centering
        \includegraphics[width=\linewidth]{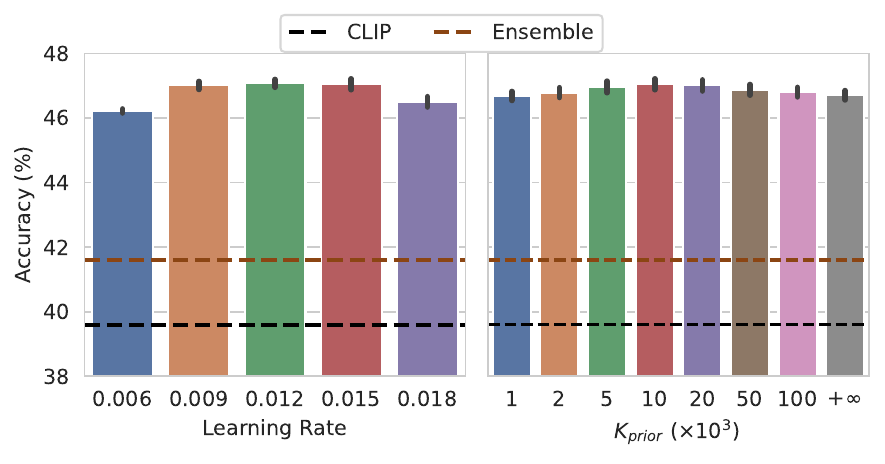}
        \vspace{-4ex}
        \caption{Hyperparameter sensitivity. }
        \label{fig:hp_sensitivity_imagenetc}
    \end{minipage}
    \hspace{0.02\linewidth}
    % \vspace{-2ex}
\end{figure}

\paragraph{Ablation study}
To understand the individual contributions of the two accumulators in {\algname}, we perform an ablation study on CIFAR-10-C comparing the full method with the following variants: (1) Mean accumulator only, which removes the gradient accumulator; (2) Gradient accumulator only, which removes the mean accumulator; and (3) No accumulators, which disables both components. We observe in Figure \ref{fig:ablation} that both accumulators contribute to the performance of {\algname}, especially under small batch sizes. The mean accumulator is essential for estimating PL-inter variance in extremely small batches, including the batch size of 1. Without it, gradients cannot be computed when the batch contains only a single class instance, rendering adaptation ineffective. Meanwhile, the gradient accumulator improves adaptation quality by reducing the noise in gradient estimates across batches. Overall, {\algname} exhibits the strongest robustness and performance when both accumulators are used, validating the necessity of their complementary roles in the online test-time adaptation setting.
Additionally, we explore adapting different layers in the visual encoder and find that updating all LayerNorm layers yields the best performance (see Appendix~\ref{appendix:exp:layer}).

%%%%%%%% %%%%%%%% %%%%%%%% %%%%%%%% %%%%%%%% %%%%%%%% %%%%%%%% %%%%%%%%

\paragraph{Hyperparameter sensitivity}
We study the sensitivity of {\algname} to its two hyperparameters: the learning rate and the prior strength $K_{\text{prior}}$, across three datasets. Results on ImageNet-C are shown in Figure \ref{fig:hp_sensitivity_imagenetc}, with results on CIFAR-10-C and CIFAR-100-C included in Appendix \ref{appendix:exp:hp}. We observe that {\algname} remains stable across a broad range of hyperparameter values, without requiring precise tuning. In particular, we find that a learning rate of 0.009 and a prior size of $K_{\text{prior}} = 10{,}000$ consistently perform well across different datasets and architectures, demonstrating the robustness and generality of the method. 

\paragraph{Additional experiments}
We further evaluate {\algname} on clean datasets (uncorrupted CIFAR-10, CIFAR-100, and ImageNet), ImageNet variants (ImageNet-A, -V2, -R, and -Sketch), and corruption benchmarks under the mixture-of-domain setting \cite{sar}. The corresponding results are provided in Appendix \ref{appendix:exp:clean}, \ref{appendix:exp:imagenet_variants}, and \ref{appendix:exp:mixture}. {\algname} demonstrates consistently strong performance across these scenarios, confirming its broad applicability.

% We study the sensitivity of Mint to its only two hyperparameters: the learning rate and the strength of the prior $K_{\text{prior}}$, across three datasets. Results on CIFAR-10-C are shown in Figure \ref{fig:hp_sensitivity_cifar10}, with CIFAR-100-C and ImageNet-C results in the Appendix \ref{appendix:exp:hp}. We find that {\algname} is robust to a wide range of hyperparameter choices. For the learning rate, performance peaks around 0.009 but remains stable between 0.006 and 0.012; extreme values (0.003 or 0.015) slightly degrade performance. For the prior strength, accuracy remains high across a broad range (from 1 to 50 ×10³), with the best results near 5 and 10 ×10³. A very large prior ($K_{\text{prior}} = \infty$) slightly reduces performance, suggesting that an overly strong prior may hinder adaptation. Overall, these results indicate that Mint performs reliably without requiring careful hyperparameter tuning.

\section{Conclusion}

In this work, we identify variance collapse in image embeddings as a key factor behind CLIP’s performance degradation under corruptions. Through theoretical analysis, we attribute this phenomenon to the image encoder encoding corruption-related patterns, which dilutes class-discriminative signals. We further show that maximizing inter-class variance, even when computed using pseudo labels, can provably enhance performance. Based on this insight, we propose Mint, a simple yet effective test-time adaptation method. Mint leverages cumulative mean and gradient accumulators to operate robustly in low-batch-size, online settings. Extensive experiments on corruption benchmarks demonstrate its strong performance and efficiency. 

\begin{ack}
This work is supported by National Science Foundation under Award No. IIS-2416070, IIS-2117902. The views and conclusions are those of the authors and should not be interpreted as representing the official policies of the funding agencies or the government.
\end{ack}

\bibliographystyle{plain}
\bibliography{main}

%%%%%%%%%%%%%%%%%%%%%%%%%%%%%%%%%%%%%%%%%%%%%%%%%%%%%%%%%%%%

\newpage
\newpage
\section*{NeurIPS Paper Checklist}

\begin{enumerate}

\item {\bf Claims}
    \item[] Question: Do the main claims made in the abstract and introduction accurately reflect the paper's contributions and scope?
    \item[] Answer: \answerYes{} % Replace by \answerYes{}, \answerNo{}, or \answerNA{}.
    \item[] Justification: In the abstract and introduction, we claim that this paper focuses on test-time adaptation of vision-language models. Contributions are clearly listed at the end of the introduction and in the abstract. 
    \item[] Guidelines:
    \begin{itemize}
        \item The answer NA means that the abstract and introduction do not include the claims made in the paper.
        \item The abstract and/or introduction should clearly state the claims made, including the contributions made in the paper and important assumptions and limitations. A No or NA answer to this question will not be perceived well by the reviewers. 
        \item The claims made should match theoretical and experimental results, and reflect how much the results can be expected to generalize to other settings. 
        \item It is fine to include aspirational goals as motivation as long as it is clear that these goals are not attained by the paper. 
    \end{itemize}

\item {\bf Limitations}
    \item[] Question: Does the paper discuss the limitations of the work performed by the authors?
    \item[] Answer: \answerYes{} % Replace by \answerYes{}, \answerNo{}, or \answerNA{}.
    \item[] Justification: Please refer to Appendix \ref{appendix:discussion:limitation}. 
    \item[] Guidelines:
    \begin{itemize}
        \item The answer NA means that the paper has no limitation while the answer No means that the paper has limitations, but those are not discussed in the paper. 
        \item The authors are encouraged to create a separate "Limitations" section in their paper.
        \item The paper should point out any strong assumptions and how robust the results are to violations of these assumptions (e.g., independence assumptions, noiseless settings, model well-specification, asymptotic approximations only holding locally). The authors should reflect on how these assumptions might be violated in practice and what the implications would be.
        \item The authors should reflect on the scope of the claims made, e.g., if the approach was only tested on a few datasets or with a few runs. In general, empirical results often depend on implicit assumptions, which should be articulated.
        \item The authors should reflect on the factors that influence the performance of the approach. For example, a facial recognition algorithm may perform poorly when image resolution is low or images are taken in low lighting. Or a speech-to-text system might not be used reliably to provide closed captions for online lectures because it fails to handle technical jargon.
        \item The authors should discuss the computational efficiency of the proposed algorithms and how they scale with dataset size.
        \item If applicable, the authors should discuss possible limitations of their approach to address problems of privacy and fairness.
        \item While the authors might fear that complete honesty about limitations might be used by reviewers as grounds for rejection, a worse outcome might be that reviewers discover limitations that aren't acknowledged in the paper. The authors should use their best judgment and recognize that individual actions in favor of transparency play an important role in developing norms that preserve the integrity of the community. Reviewers will be specifically instructed to not penalize honesty concerning limitations.
    \end{itemize}

\item {\bf Theory assumptions and proofs}
    \item[] Question: For each theoretical result, does the paper provide the full set of assumptions and a complete (and correct) proof?
    \item[] Answer: \answerYes{}% Replace by \answerYes{}, \answerNo{}, or \answerNA{}.
    \item[] Justification: We clearly state the main assumptions in Section \ref{sec:analysis}. Formal statements and proofs are given in Appendix \ref{appendix:analysis}. 
    \item[] Guidelines:
    \begin{itemize}
        \item The answer NA means that the paper does not include theoretical results. 
        \item All the theorems, formulas, and proofs in the paper should be numbered and cross-referenced.
        \item All assumptions should be clearly stated or referenced in the statement of any theorems.
        \item The proofs can either appear in the main paper or the supplemental material, but if they appear in the supplemental material, the authors are encouraged to provide a short proof sketch to provide intuition. 
        \item Inversely, any informal proof provided in the core of the paper should be complemented by formal proofs provided in appendix or supplemental material.
        \item Theorems and Lemmas that the proof relies upon should be properly referenced. 
    \end{itemize}

    \item {\bf Experimental result reproducibility}
    \item[] Question: Does the paper fully disclose all the information needed to reproduce the main experimental results of the paper to the extent that it affects the main claims and/or conclusions of the paper (regardless of whether the code and data are provided or not)?
    \item[] Answer: \answerYes{} % Replace by \answerYes{}, \answerNo{}, or \answerNA{}.
    \item[] Justification: In the paper, we clearly specified the methods we used to obtain the experimental results and all the hyperparameters used in the process, which can fully support the reproducibility of the experiment. More details are provided in Appendix \ref{appendix:exp:detail}. 
    \item[] Guidelines: 
    \begin{itemize}
        \item The answer NA means that the paper does not include experiments.
        \item If the paper includes experiments, a No answer to this question will not be perceived well by the reviewers: Making the paper reproducible is important, regardless of whether the code and data are provided or not.
        \item If the contribution is a dataset and/or model, the authors should describe the steps taken to make their results reproducible or verifiable. 
        \item Depending on the contribution, reproducibility can be accomplished in various ways. For example, if the contribution is a novel architecture, describing the architecture fully might suffice, or if the contribution is a specific model and empirical evaluation, it may be necessary to either make it possible for others to replicate the model with the same dataset, or provide access to the model. In general. releasing code and data is often one good way to accomplish this, but reproducibility can also be provided via detailed instructions for how to replicate the results, access to a hosted model (e.g., in the case of a large language model), releasing of a model checkpoint, or other means that are appropriate to the research performed.
        \item While NeurIPS does not require releasing code, the conference does require all submissions to provide some reasonable avenue for reproducibility, which may depend on the nature of the contribution. For example
        \begin{enumerate}
            \item If the contribution is primarily a new algorithm, the paper should make it clear how to reproduce that algorithm.
            \item If the contribution is primarily a new model architecture, the paper should describe the architecture clearly and fully.
            \item If the contribution is a new model (e.g., a large language model), then there should either be a way to access this model for reproducing the results or a way to reproduce the model (e.g., with an open-source dataset or instructions for how to construct the dataset).
            \item We recognize that reproducibility may be tricky in some cases, in which case authors are welcome to describe the particular way they provide for reproducibility. In the case of closed-source models, it may be that access to the model is limited in some way (e.g., to registered users), but it should be possible for other researchers to have some path to reproducing or verifying the results.
        \end{enumerate}
    \end{itemize}

\item {\bf Open access to data and code}
    \item[] Question: Does the paper provide open access to the data and code, with sufficient instructions to faithfully reproduce the main experimental results, as described in supplemental material?
    \item[] Answer: \answerYes{} % Replace by \answerYes{}, \answerNo{}, or \answerNA{}.
    \item[] Justification: We use open-source datasets, and provide the code in the supplemental material. 
    \item[] Guidelines:
    \begin{itemize}
        \item The answer NA means that paper does not include experiments requiring code.
        \item Please see the NeurIPS code and data submission guidelines (\url{https://nips.cc/public/guides/CodeSubmissionPolicy}) for more details.
        \item While we encourage the release of code and data, we understand that this might not be possible, so “No” is an acceptable answer. Papers cannot be rejected simply for not including code, unless this is central to the contribution (e.g., for a new open-source benchmark).
        \item The instructions should contain the exact command and environment needed to run to reproduce the results. See the NeurIPS code and data submission guidelines (\url{https://nips.cc/public/guides/CodeSubmissionPolicy}) for more details.
        \item The authors should provide instructions on data access and preparation, including how to access the raw data, preprocessed data, intermediate data, and generated data, etc.
        \item The authors should provide scripts to reproduce all experimental results for the new proposed method and baselines. If only a subset of experiments are reproducible, they should state which ones are omitted from the script and why.
        \item At submission time, to preserve anonymity, the authors should release anonymized versions (if applicable).
        \item Providing as much information as possible in supplemental material (appended to the paper) is recommended, but including URLs to data and code is permitted.
    \end{itemize}

\item {\bf Experimental setting/details}
    \item[] Question: Does the paper specify all the training and test details (e.g., data splits, hyperparameters, how they were chosen, type of optimizer, etc.) necessary to understand the results?
    \item[] Answer: \answerYes{} % Replace by \answerYes{}, \answerNo{}, or \answerNA{}.
    \item[] Justification: We provide key information in Section \ref{sec:experiment}, and other details in Appendix \ref{appendix:exp:detail}. 
    \item[] Guidelines:
    \begin{itemize}
        \item The answer NA means that the paper does not include experiments.
        \item The experimental setting should be presented in the core of the paper to a level of detail that is necessary to appreciate the results and make sense of them.
        \item The full details can be provided either with the code, in appendix, or as supplemental material.
    \end{itemize}

\item {\bf Experiment statistical significance}
    \item[] Question: Does the paper report error bars suitably and correctly defined or other appropriate information about the statistical significance of the experiments?
    \item[] Answer: \answerYes{} % Replace by \answerYes{}, \answerNo{}, or \answerNA{}.
    \item[] Justification: We report results with mean and standard deviation from five independent runs with different random seeds. Notice that for space limits, we provide the error bar of Table \ref{tab:acc:main} in Appendix \ref{appendix:exp:rq1} instead. 
    \item[] Guidelines:
    \begin{itemize}
        \item The answer NA means that the paper does not include experiments.
        \item The authors should answer "Yes" if the results are accompanied by error bars, confidence intervals, or statistical significance tests, at least for the experiments that support the main claims of the paper.
        \item The factors of variability that the error bars are capturing should be clearly stated (for example, train/test split, initialization, random drawing of some parameter, or overall run with given experimental conditions).
        \item The method for calculating the error bars should be explained (closed form formula, call to a library function, bootstrap, etc.)
        \item The assumptions made should be given (e.g., Normally distributed errors).
        \item It should be clear whether the error bar is the standard deviation or the standard error of the mean.
        \item It is OK to report 1-sigma error bars, but one should state it. The authors should preferably report a 2-sigma error bar than state that they have a 96\% CI, if the hypothesis of Normality of errors is not verified.
        \item For asymmetric distributions, the authors should be careful not to show in tables or figures symmetric error bars that would yield results that are out of range (e.g. negative error rates).
        \item If error bars are reported in tables or plots, The authors should explain in the text how they were calculated and reference the corresponding figures or tables in the text.
    \end{itemize}

\item {\bf Experiments compute resources}
    \item[] Question: For each experiment, does the paper provide sufficient information on the computer resources (type of compute workers, memory, time of execution) needed to reproduce the experiments?
    \item[] Answer: \answerYes{} % Replace by \answerYes{}, \answerNo{}, or \answerNA{}. 
    \item[] Justification: Please refer to Appendix \ref{appendix:exp:resource}. 
    \item[] Guidelines:
    \begin{itemize}
        \item The answer NA means that the paper does not include experiments.
        \item The paper should indicate the type of compute workers CPU or GPU, internal cluster, or cloud provider, including relevant memory and storage.
        \item The paper should provide the amount of compute required for each of the individual experimental runs as well as estimate the total compute. 
        \item The paper should disclose whether the full research project required more compute than the experiments reported in the paper (e.g., preliminary or failed experiments that didn't make it into the paper). 
    \end{itemize}
    
\item {\bf Code of ethics}
    \item[] Question: Does the research conducted in the paper conform, in every respect, with the NeurIPS Code of Ethics \url{https://neurips.cc/public/EthicsGuidelines}?
    \item[] Answer: \answerYes{} % Replace by \answerYes{}, \answerNo{}, or \answerNA{}.
    \item[] Justification: We strictly adhere to the NeurIPS Code of Ethics.
    \item[] Guidelines:
    \begin{itemize}
        \item The answer NA means that the authors have not reviewed the NeurIPS Code of Ethics.
        \item If the authors answer No, they should explain the special circumstances that require a deviation from the Code of Ethics.
        \item The authors should make sure to preserve anonymity (e.g., if there is a special consideration due to laws or regulations in their jurisdiction).
    \end{itemize}

\item {\bf Broader impacts}
    \item[] Question: Does the paper discuss both potential positive societal impacts and negative societal impacts of the work performed?
    \item[] Answer: \answerYes{} % Replace by \answerYes{}, \answerNo{}, or \answerNA{}.
    \item[] Justification: Please refer to Appendix \ref{appendix:discussion:impact}. 
    \item[] Guidelines:
    \begin{itemize}
        \item The answer NA means that there is no societal impact of the work performed.
        \item If the authors answer NA or No, they should explain why their work has no societal impact or why the paper does not address societal impact.
        \item Examples of negative societal impacts include potential malicious or unintended uses (e.g., disinformation, generating fake profiles, surveillance), fairness considerations (e.g., deployment of technologies that could make decisions that unfairly impact specific groups), privacy considerations, and security considerations.
        \item The conference expects that many papers will be foundational research and not tied to particular applications, let alone deployments. However, if there is a direct path to any negative applications, the authors should point it out. For example, it is legitimate to point out that an improvement in the quality of generative models could be used to generate deepfakes for disinformation. On the other hand, it is not needed to point out that a generic algorithm for optimizing neural networks could enable people to train models that generate Deepfakes faster.
        \item The authors should consider possible harms that could arise when the technology is being used as intended and functioning correctly, harms that could arise when the technology is being used as intended but gives incorrect results, and harms following from (intentional or unintentional) misuse of the technology.
        \item If there are negative societal impacts, the authors could also discuss possible mitigation strategies (e.g., gated release of models, providing defenses in addition to attacks, mechanisms for monitoring misuse, mechanisms to monitor how a system learns from feedback over time, improving the efficiency and accessibility of ML).
    \end{itemize}
    
\item {\bf Safeguards}
    \item[] Question: Does the paper describe safeguards that have been put in place for responsible release of data or models that have a high risk for misuse (e.g., pretrained language models, image generators, or scraped datasets)?
    \item[] Answer: \answerNA{} % Replace by \answerYes{}, \answerNo{}, or \answerNA{}.
    \item[] Justification: Our paper does not release new data or models. 
    \item[] Guidelines:
    \begin{itemize}
        \item The answer NA means that the paper poses no such risks.
        \item Released models that have a high risk for misuse or dual-use should be released with necessary safeguards to allow for controlled use of the model, for example by requiring that users adhere to usage guidelines or restrictions to access the model or implementing safety filters. 
        \item Datasets that have been scraped from the Internet could pose safety risks. The authors should describe how they avoided releasing unsafe images.
        \item We recognize that providing effective safeguards is challenging, and many papers do not require this, but we encourage authors to take this into account and make a best faith effort.
    \end{itemize}

\item {\bf Licenses for existing assets}
    \item[] Question: Are the creators or original owners of assets (e.g., code, data, models), used in the paper, properly credited and are the license and terms of use explicitly mentioned and properly respected?
    \item[] Answer: \answerYes{} % Replace by \answerYes{}, \answerNo{}, or \answerNA{}.
    \item[] Justification: All assets used in the paper are properly credited, and their licenses and terms of use have been explicitly mentioned and respected if provided in the original paper. 
    \item[] Guidelines:
    \begin{itemize}
        \item The answer NA means that the paper does not use existing assets.
        \item The authors should cite the original paper that produced the code package or dataset.
        \item The authors should state which version of the asset is used and, if possible, include a URL.
        \item The name of the license (e.g., CC-BY 4.0) should be included for each asset.
        \item For scraped data from a particular source (e.g., website), the copyright and terms of service of that source should be provided.
        \item If assets are released, the license, copyright information, and terms of use in the package should be provided. For popular datasets, \url{paperswithcode.com/datasets} has curated licenses for some datasets. Their licensing guide can help determine the license of a dataset.
        \item For existing datasets that are re-packaged, both the original license and the license of the derived asset (if it has changed) should be provided.
        \item If this information is not available online, the authors are encouraged to reach out to the asset's creators.
    \end{itemize}

\item {\bf New assets}
    \item[] Question: Are new assets introduced in the paper well documented and is the documentation provided alongside the assets?
    \item[] Answer: \answerNA{} % Replace by \answerYes{}, \answerNo{}, or \answerNA{}.
    \item[] Justification: Our paper does not release new assets.
    \item[] Guidelines:
    \begin{itemize}
        \item The answer NA means that the paper does not release new assets.
        \item Researchers should communicate the details of the dataset/code/model as part of their submissions via structured templates. This includes details about training, license, limitations, etc. 
        \item The paper should discuss whether and how consent was obtained from people whose asset is used.
        \item At submission time, remember to anonymize your assets (if applicable). You can either create an anonymized URL or include an anonymized zip file.
    \end{itemize}

\item {\bf Crowdsourcing and research with human subjects}
    \item[] Question: For crowdsourcing experiments and research with human subjects, does the paper include the full text of instructions given to participants and screenshots, if applicable, as well as details about compensation (if any)? 
    \item[] Answer: \answerNA{} % Replace by \answerYes{}, \answerNo{}, or \answerNA{}.
    \item[] Justification: Our paper does not involve crowdsourcing nor research with human subjects.
    \item[] Guidelines:
    \begin{itemize}
        \item The answer NA means that the paper does not involve crowdsourcing nor research with human subjects.
        \item Including this information in the supplemental material is fine, but if the main contribution of the paper involves human subjects, then as much detail as possible should be included in the main paper. 
        \item According to the NeurIPS Code of Ethics, workers involved in data collection, curation, or other labor should be paid at least the minimum wage in the country of the data collector. 
    \end{itemize}

\item {\bf Institutional review board (IRB) approvals or equivalent for research with human subjects}
    \item[] Question: Does the paper describe potential risks incurred by study participants, whether such risks were disclosed to the subjects, and whether Institutional Review Board (IRB) approvals (or an equivalent approval/review based on the requirements of your country or institution) were obtained?
    \item[] Answer: \answerNA{} % Replace by \answerYes{}, \answerNo{}, or \answerNA{}.
    \item[] Justification: Our paper does not involve crowdsourcing nor research with human subjects.
    \item[] Guidelines:
    \begin{itemize}
        \item The answer NA means that the paper does not involve crowdsourcing nor research with human subjects.
        \item Depending on the country in which research is conducted, IRB approval (or equivalent) may be required for any human subjects research. If you obtained IRB approval, you should clearly state this in the paper. 
        \item We recognize that the procedures for this may vary significantly between institutions and locations, and we expect authors to adhere to the NeurIPS Code of Ethics and the guidelines for their institution. 
        \item For initial submissions, do not include any information that would break anonymity (if applicable), such as the institution conducting the review.
    \end{itemize}

\item {\bf Declaration of LLM usage}
    \item[] Question: Does the paper describe the usage of LLMs if it is an important, original, or non-standard component of the core methods in this research? Note that if the LLM is used only for writing, editing, or formatting purposes and does not impact the core methodology, scientific rigorousness, or originality of the research, declaration is not required.
    %this research? 
    \item[] Answer: \answerNA{} % Replace by \answerYes{}, \answerNo{}, or \answerNA{}.
    \item[] Justification: LLM is used only for writing, editing, or formatting purposes in this paper. 
    \item[] Guidelines:
    \begin{itemize}
        \item The answer NA means that the core method development in this research does not involve LLMs as any important, original, or non-standard components.
        \item Please refer to our LLM policy (\url{https://neurips.cc/Conferences/2025/LLM}) for what should or should not be described.
    \end{itemize}

\end{enumerate}

%%%%%%%%%%%%%%%%%%%%%%%%%%%%%%%%%%%%%%%%%%%%%%%%%%%%%%%%%%%%

\newpage
\appendix

\section*{Appendix}

\vspace{20pt}

\addtocontents{toc}{\protect\setcounter{tocdepth}{2}}

\tableofcontents

\newpage
\section{Discussion}

\subsection{Additional related works} \label{appendix:discussion:related_works}

In this subsection, we discuss additional related work on general test-time adaptation. Many of these methods have inspired recent advances in TTA algorithms for VLMs.

\paragraph{Generic test-time adaptation (TTA)}
Most TTA methods aim to improve model accuracy by optimizing a carefully designed unsupervised loss on unlabeled test data. A prominent line of work minimizes the entropy of model predictions, based on the intuition that entropy quantifies prediction uncertainty. Pioneered by Tent \cite{tent}, these methods typically update the running statistics and affine parameters of batch normalization \cite{batchnorm} layers. However, entropy minimization is often unstable, and many subsequent works \cite{eata,sar,deyo} focus on improving its robustness. One important variant is marginal entropy \cite{memo}, which captures a model’s uncertainty across different augmentations of the same input. This idea has inspired several follow-up TTA approaches \cite{tpt,tps} for VLMs.

Another line of potential approaches focuses on restoring uncorrupted images from corrupted ones, using generative techniques such as diffusion models \cite{dda} or super-resolution \cite{lrfm,robustsam}. These methods do not require adapting the model at test time. However, as noted in \cite{sar}, they often perform well on certain types of corruption but poorly on others, indicating limited generalization across corruption types.

\subsection{Limitations} \label{appendix:discussion:limitation}

While our analysis reveals a consistent variance collapse pattern across multiple datasets and corruption types, it primarily focuses on natural distribution shifts and classification tasks. Extending our analysis and algorithm to broader types of distribution shifts (e.g., adversarial perturbations) and more diverse tasks (e.g., object detection, semantic segmentation) represents an important direction for future work.

\subsection{Broader impacts}  \label{appendix:discussion:impact}

Our work focuses on understanding and mitigating the degradation of vision-language models under distribution shift, particularly in the context of image corruption. On the positive side, improving model robustness can enhance the reliability of real-world applications such as accessibility tools, autonomous systems, and content moderation, especially under suboptimal conditions. By providing theoretical insights and simple, efficient test-time adaptation methods, our work contributes toward safer and more dependable AI deployments.

We do not anticipate significant negative societal impacts. Our method is unsupervised, operates solely at test time, and does not require access to sensitive data or any form of user interaction. Nonetheless, as with all performance-enhancing techniques, there is potential for misuse in contexts where robustness could amplify existing biases or be deployed without appropriate oversight. We encourage future work to consider fairness and accountability as these methods are applied more broadly.
\newpage
\section{Theoretical analysis} \label{appendix:analysis}

\subsection{Variance decomposition} \label{appendix:analysis:var_decompose}

In this section we give formal proof of variance decomposition. 

\begin{lemma}
    \begin{align}
        \gVGTtotal = \gVGTintra + \gVGTinter, \quad \gVPLtotal = \gVPLintra + \gVPLinter. 
    \end{align}
\end{lemma}

\begin{proof}
We define ``prior'' for each class $c = 1, \cdots, C$: 
\begin{align}
    \bar{y}_c = \frac{1}{N} \sum_{i=1}^N y_{ic},
\end{align}
The class means are
\begin{align}
    \bar\vz_c := \frac{\sum_{i=1}^N y_{ic} \cdot \vz_i}{\sum_{i=1}^N y_{ic}} = \frac{1}{N} \sum_{i=1}^N \frac{y_{ic}}{\bar{y}_c} \cdot \vz_i. 
    \label{eq:class_mean}
\end{align}
The global mean is
\begin{align}
    \bar\vz = \frac{1}{N} \sum_{i=1}^N \vz_i. 
\end{align}
Notice that for all class $c = 1, \cdots, C$, we have
\begin{align*}
    \frac{1}{N} \sum_{i=1}^N \frac{y_{ic}}{\bar{y}_c} (\vz_i - \bar\vz_c) 
    &= \left( \frac{1}{N} \sum_{i=1}^N \frac{y_{ic}}{\bar{y}_c} \vz_i \right) - \left( \frac{1}{N} \sum_{i=1}^N \frac{y_{ic}}{\bar{y}_c} \bar\vz_c \right) \\
    &= \bar\vz_c - \left( \frac{1}{N} \sum_{i=1}^N \frac{y_{ic}}{\bar{y}_c} \bar\vz_c \right) \tag{definition of $\bar\vz_c$} \\
    &= \bar\vz_c - \bar\vz_c \tag{definition of $\bar{y}_c$} \\
    &= \vzero
\end{align*}

Therefore, 
\begin{align*}
    \gVGTtotal
    &= \frac{1}{N \cdot C} \sum_{i=1}^N \sum_{c=1}^C \frac{y_{ic}}{\bar{y}_c} \left\| \vz_i - \bar\vz \right\|_2^2 \\
    &= \frac{1}{N \cdot C} \sum_{i=1}^N \sum_{c=1}^C \frac{y_{ic}}{\bar{y}_c} \left\| \vz_i - \bar\vz_c + \bar\vz_c - \bar\vz \right\|_2^2 \\
    &= \frac{1}{N \cdot C} \sum_{i=1}^N \sum_{c=1}^C \frac{y_{ic}}{\bar{y}_c} \left( \left\| \vz_i - \bar\vz_c \right\|_2^2 + \left\| \bar\vz_c - \bar\vz \right\|_2^2 + 2 \left( \vz_i - \bar\vz_c \right)^\top \left( \bar\vz_c - \bar\vz  \right) \right) \\
    &= \frac{1}{N \cdot C} \sum_{i=1}^N \sum_{c=1}^C \frac{y_{ic}}{\bar{y}_c} \left\| \vz_i - \bar\vz_c \right\|_2^2 + \frac{1}{N \cdot C} \sum_{i=1}^N \sum_{c=1}^C \frac{y_{ic}}{\bar{y}_c} \left\| \bar\vz_c - \bar\vz \right\|_2^2 \\
    &= \frac{1}{N \cdot C} \sum_{i=1}^N \sum_{c=1}^C \frac{y_{ic}}{\bar{y}_c} \left\| \vz_i - \bar\vz_c \right\|_2^2 + \frac{1}{C} \sum_{c=1}^C \left\| \bar\vz_c - \bar\vz \right\|_2^2 \tag{definition of $\bar{y}_c$} \\
    &= \gVGTintra + \gVGTinter
\end{align*}

By replacing each $y_{ic}$ with $\hat{y}_{ic}$, $\bar{\vz}_c$ with $\tilde{\vz}_c$, and $\bar{\vz}$ with $\tilde{\vz}$, and repeating the above steps, it is straightforward to prove $\gVPLtotal = \gVPLintra + \gVPLinter$. 

\end{proof}

\newpage
\subsection{Theoretical setup}

This section introduces the setup and assumptions of our theoretical analysis. For simplicity, we focus on a binary classification setting where $C=2$. While the standard notation of label for image $i$ is $\vy_i = [y_{i0}, y_{i1}]^\top \in \R^2$, we write $y_i = y_{i1}$ for brevity, with a mild abuse of notation. We also assume there is no label imbalance, i.e., $\Pr(y_i = 0) = \Pr(y_1=1) = \frac{1}{2}$. 

\paragraph{Image latent representation}
Motivated by \cite{shift_analysis}, we assume that each image can be mapped to a disentangled latent representation $\vv_i = [\vvcls_i; \vvirr_i; \vvshift_i; \vvnoise_i] \in \R^d$, composed of four components: 
\begin{enumerate}
    \item \textit{Class-relevant feature} $\vvcls_i \in \R^{\dcls}$: 
        Semantic feature that are directly predictive of the class label, 
        $\vvcls_i = \vmu$ for $y_i = 1$ and $\vvcls_i = -\vmu$ for $y_i = 0$. 
        
    \item \textit{Class-irrelevant feature} $\vvirr_i \in \R^{\dirr}$: 
        Features that are unrelated to the classification task, such as background information. It is preserved during pretraining due to CLIP's general representation learning objective. We assume $\vvirr_i \sim \text{Rademacher}^{\dirr}$, i.e., uniformly distributed in $\{-1, 1\}^{\dirr}$. 
        
    \item \textit{Structured distribution shift} $\vvshift_i \in \R^{\dshift}$: 
        Features representing systematic distribution changes in the target domain, such as weather conditions or digital transforms. We assume $\vvshift_i = s \cdot \vdelta$, where $s$ indicates the severity of corruption or distribution shift. 
        
    \item \textit{Unstructured noise} $\vvnoise_i$: 
        Random noise introduced by the corruption process. We assume $\vvnoise \sim s \cdot \text{Rademacher}^{\dnoise}$, i.e., uniformly distributed in $\{-1, 1\}^{\dnoise}$. 
\end{enumerate}

Notice that by controlling the ratio of $s, \| \vmu \|_2, \| \vdelta \|_2$, we can freely adjust the ratio for four components. 

% Notice that by controlling the ratio of $s, \| \vmu \|_2, \| \vdelta \|_2$, we can freely adjust the ratio for each component. 
% Following the structure of CLIP’s visual encoder, we assume that the latent representation $\vv$ first passes through a LayerNorm layer~\cite{layernorm} with a linear transformation, followed by normalization to unit length. For analytical simplicity, we omit the demeaning step in LayerNorm and ignore the bias term in its parameters,\footnote{This simplification is also known as RMSNorm~\cite{rmsnorm}.} under which the image embedding can be formulated as
% \begin{align}
%     \vz_i = \normalize \left( \frac{\vv_i}{\sqrt{\Var[\vv_i]}} \odot \vw \right), 
% \end{align}
% where $\vw \in \R^{d}$ is the learnable weight of the LayerNorm layer, $\normalize(\cdot)$ denotes $\ell_2$ normalization. For simplicity, we assume $\vw = \vone$ at initialization. 

\paragraph{LayerNorm and image embedding}
Following the structure of CLIP's visual encoder, we assume that the latent representation $\vv_i$ first passes through a LayerNorm layer \cite{layernorm} with linear transformation, and then normalized to unit length. For analytical simplicity, we omit the demeaning step LayerNorm and ignore the bias term in its parameters. This simplification is also known as RMSNorm~\cite{rmsnorm}. Under this simplification, the image embedding can be expressed as
\begin{align}
    \vz_i = \normalize \left( \frac{\vv_i}{\sqrt{\Var[\vv_i]}} \odot \vw \right), 
\end{align}
where $\odot$ represents element-wise multiplication of vectors, $\vw = [\vwcls; \vwirr; \vwshift; \vwnoise] \in \R^d$ is the LayerNorm weights, and $\normalize(\cdot)$ denotes $\ell_2$ normalization. For simplicity, we assume $\vw = \vone$ at initialization. $\vw$ is updated during TTA. Since $\sqrt{\Var[\vv_i]}$ is just a scalar, the equation above can be further reduced to
\begin{align}
    \vz_i = \normalize\left(\vv_i \odot \vw \right) = \frac{\vv_i \odot \vw}{\| \vv_i \odot \vw \|_2}. 
\end{align}

\paragraph{Text embedding and prediction}
Let $\vt_0, \vt_1$ denotes the text embedding for class 0 and 1, respectively. The model prediction is given by
\begin{align}
    y_i = \begin{cases}
        0, & \text{when }\vz_i^\top \vt_0 \geq \vz_i^\top \vt_1 \\
        1, & \text{when }\vz_i^\top \vt_0 < \vz_i^\top \vt_1 \\
    \end{cases}
\end{align}

\newpage
\subsection{Change of variances under corruption}

This section studies the behavior of various types of variance with increasing corruption severity $s$. 

\var*

\begin{proof}
    We first compute the normalizing factor for each image: 
    \begin{align*}
        \| \vv_i \odot \vw \|_2^2
        &= \| \vvcls_i \odot \vwcls \|_2^2 + \| \vvirr_i \odot \vwirr \|_2^2 + \| \vvshift_i \odot \vwshift \|_2^2 + \| \vvnoise_i \odot \vwnoise \|_2^2 \\
        &= \| \vmu \odot \vwcls \|_2^2 + \| \vwirr \|_2^2 + s^2 \cdot \| \vdelta \odot \vwshift \|_2^2 + s^2 \cdot \| \vwnoise \|_2^2 \\
        &= \| \vmu \|_2^2 + \dirr + s^2 \cdot (\| \vdelta \|_2^2 +  \dnoise) \tag{at initialization $\vw = \vone$}
    \end{align*}
    Notice that this normalizing factor is the same for each image, and is a function of $\vw$ and severity $s$. Let 
    \begin{align*}
        Z(\vw, s) = \sqrt{\| \vmu \odot \vwcls \|_2^2 + \| \vwirr \|_2^2 + s^2 \cdot \| \vdelta \odot \vwshift \|_2^2 + s^2 \cdot \| \vwnoise \|_2^2}
    \end{align*}
    denote the normalizing factor. Under infinite sample size, the total mean $\bar{\vz}$ and class means $\bar{\vz}_0, \bar{\vz}_1$ can be expressed as: 
    \begin{align*}
        \bar{\vz} &\xrightarrow{p} \E \vz_i = \frac{1}{Z(\vw, s)} \cdot [\vzero; \vzero; s \cdot \vdelta \odot \vwshift; \vzero] \\
        \bar{\vz}_0 &\xrightarrow{p} \E [ \vz_i | y_i = 0] = \frac{1}{Z(\vw, s)} \cdot [\vmu \odot \vwcls; \vzero; s \cdot \vdelta \odot \vwshift; \vzero] \\
        \bar{\vz}_1 &\xrightarrow{p} \E [ \vz_i | y_i = 1] = \frac{1}{Z(\vw, s)} \cdot [-\vmu \odot \vwcls; \vzero; s \cdot \vdelta \odot \vwshift; \vzero] 
    \end{align*}
    The GT-total variance: 
    \begin{align*}
        \gVGTtotal &\xrightarrow{p} \E \| \vz_i - \E \vz_i \|_2^2 \\
        &= \frac{1}{Z(\vw, s)^2} \cdot \left( \| \vmu \odot \vwcls \|_2^2 + \| \vwirr \|_2^2 + 0 + \| \vwnoise \|_2^2 \right) \\
        &= \frac{\| \vmu \odot \vwcls \|_2^2 + \| \vwirr \|_2^2 + s^2 \cdot \| \vwnoise \|_2^2}{\| \vmu \odot \vwcls \|_2^2 + \| \vwirr \|_2^2 + s^2 \cdot \| \vdelta \odot \vwshift \|_2^2 + s^2 \cdot \| \vwnoise \|_2^2} \\
        &= \frac{\| \vmu \|_2^2 + \dirr + s^2 \cdot \dnoise}{\| \vmu \|_2^2 + \dirr + s^2 \cdot ( \|\vdelta \|_2^2 + \dnoise)} \tag{at initialization $\vw = \vone$}
    \end{align*}
    The GT-inter variance: 
    \begin{align*}
        \gVGTinter &\xrightarrow{p} \frac{1}{2} \sum_{c=1}^2  \left\| \E [\vz_i | y_i = c] - \E \vz_i \right\|_2^2 \\
        &= \frac{1}{Z(\vw, s)^2} \cdot \| \vmu \|_2^2 \\
        &= \frac{\| \vmu \odot \vwcls \|_2^2}{\| \vmu \odot \vwcls \|_2^2 + \| \vwirr \|_2^2 + s^2 \cdot \| \vdelta \odot \vwshift \|_2^2 + s^2 \cdot \| \vwnoise \|_2^2} \\
        &= \frac{\| \vmu \|_2^2 }{\| \vmu \|_2^2 + \dirr + s^2 \cdot ( \|\vdelta \|_2^2 + \dnoise)} \tag{at initialization $\vw = \vone$}
    \end{align*}
    And the GT-intra variance: 
    \begin{align*}
        \gVGTintra &= \gVGTtotal - \gVGTinter \\
        &\xrightarrow{p} \frac{\| \vwirr \|_2^2 + s^2 \cdot \| \vwnoise \|_2^2}{\| \vmu \odot \vwcls \|_2^2 + \| \vwirr \|_2^2 + s^2 \cdot \| \vdelta \odot \vwshift \|_2^2 + s^2 \cdot \| \vwnoise \|_2^2} \\
        &= \frac{\dirr + s^2 \cdot \dnoise}{\| \vmu \|_2^2 + \dirr + s^2 \cdot ( \|\vdelta \|_2^2 + \dnoise)} \tag{at initialization $\vw = \vone$} 
    \end{align*}    
\end{proof}
\newpage
\subsection{Adaptation}

In this section, we derive how maximizing the pseudo-label inter-class (PL-inter) variance influences the learned representation, when only the LayerNorm parameters are updated during test-time adaptation. 

\begin{lemma}
    When the sample size $N \to +\infty$, 
    \begin{align*}
        \gVPLinter \xrightarrow{p} \frac{1}{2} \left(\frac{1}{(\E \hat{y}_i)^2} + \frac{1}{(1 - \E \hat{y}_i)^2} \right) \cdot \| \Cov(\vz_i, \hat{y}_i) \|_2^2. 
    \end{align*}
\end{lemma}
\begin{proof}
    \begin{align*}
        \E[\vz_i | \hat{y}_i = 1] &= \frac{\E[\hat{y}_i \cdot \vz_i]}{\E \hat{y}_i} = \frac{\E\hat{y}_i \cdot \E\vz_i + \Cov(\vz_i, \hat{y}_i)}{\E \hat{y}_i} = \E\vz_i + \frac{\Cov(\vz_i, \hat{y}_i)}{\E \hat{y}_i} \\
        \E[\vz_i | \hat{y}_i = 0] &= \E\vz_i - \frac{\Cov(\vz_i, \hat{y}_i)}{1 - \E \hat{y}_i} \\
        \gVPLinter &\xrightarrow{p} \frac{1}{2} \sum_{c=1}^2  \left\| \E [\vz_i | \hat{y}_i = c] - \E \vz_i \right\|_2^2  = \frac{1}{2} \left(\frac{1}{(\E \hat{y}_i)^2} + \frac{1}{(1 - \E \hat{y}_i)^2} \right) \cdot \| \Cov(\vz_i, \hat{y}_i) \|_2^2
    \end{align*}
    
\end{proof}

\begin{remark}
    \begin{align*}
        \Cov(\vz_i, \hat{y}_i) &\sim \Cov(\vz_i, \vz_i^\top (\vt_1 - \vt_0)) = \Sigma_{\vz_i} (\vt_1 - \vt_0)
    \end{align*}
    where $\Sigma_{\vz_i} $ is the covariance matrix of $\vz_i$ and $\vt_0, \vt_1$ are the text embedding of class 0 and 1. This implies that maximizing PL-inter will enhance those features that (1) have high variance, and (2) are more relevant to the classification task described by the text embedding. 
\end{remark}

\adapt*

\begin{proof}
    Similar to the procedure of deriving GT-inter, we start by computing the total mean $\tilde\vz$ and pseudo-class means $\tilde\vz_0, \tilde\vz_1$. 
    \begin{align*}
        \tilde\vz &\xrightarrow{p} \E \vz_i = \frac{1}{Z(\vw, s)} \cdot [\vzero; \vzero; s \cdot \vdelta \odot \vwshift; \vzero] \\
        \tilde{\vz}_1 &\xrightarrow{p} \E [ \vz_i | \hat{y}_i = 1] 
        = \frac{1}{\E \hat{y}_i} \cdot \frac{1}{Z(\vw, s)} \cdot \E [\hat{y}_i \cdot \vv_i \odot \vw] \\
        \tilde{\vz}_0 &\xrightarrow{p} \E [ \vz_i | \hat{y}_i = 0] 
        = \frac{1}{1 - \E \hat{y}_i} \cdot \frac{1}{Z(\vw, s)} \cdot \E [(1 - \hat{y}_i) \cdot \vv_i \odot \vw]
    \end{align*}
    where $Z(\vw, s) = \sqrt{\| \vmu \odot \vwcls \|_2^2 + \| \vwirr \|_2^2 + s^2 \cdot \| \vdelta \odot \vwshift \|_2^2 + s^2 \cdot \| \vwnoise \|_2^2}$ is the normalizing factor we defined in the proof of Theorem \ref{thm:var}. For four components of the feature: % \bao{Could be more detailed here}
    \begin{align*}
        \E[\hat{y}_i \cdot \vvcls_i \odot \vwcls] &= \E[\hat{y}_i \cdot (2 y_i - 1) \cdot \vmu  \odot \vwcls] = 2 \Cov (\hat{y}_i, y_i) \cdot \vmu  \odot \vwcls \\
        \E[\hat{y}_i \cdot \vvirr_i \odot \vwirr] &= \Cov(\hat{y}_i, \vvirr) \odot \vwirr \\
        \E[\hat{y}_i \cdot \vvshift_i \odot \vwshift] &= \E \hat{y}_i \cdot s \cdot \vdelta \odot \vwshift \\
        \E[\hat{y}_i \cdot \vvnoise_i \odot \vwnoise] &= \Cov(\hat{y}_i, \vvnoise) \odot \vwnoise
    \end{align*}
    And similarly, 
    \begin{align*}
        \E[(1 - \hat{y}_i) \cdot \vvcls_i \odot \vwcls] &= \E[(1 - \hat{y}_i) \cdot (2 y_i - 1) \cdot \vmu  \odot \vwcls] = - 2 \Cov (\hat{y}_i, y_i) \cdot \vmu  \odot \vwcls \\
        \E[(1 - \hat{y}_i)\cdot \vvirr_i \odot \vwirr] &= - \Cov(\hat{y}_i, \vvirr) \odot \vwirr \\
        \E[(1 - \hat{y}_i) \cdot \vvshift_i \odot \vwshift] &= \E (1 - \hat{y}_i) \cdot s \cdot \vdelta \odot \vwshift \\
        \E[(1 - \hat{y}_i) \cdot \vvnoise_i \odot \vwnoise] &= - \Cov(\hat{y}_i, \vvnoise) \odot \vwnoise 
    \end{align*}
    Therefore, we have
    \begin{align*}
        \gVPLinter &\xrightarrow{p} \frac{1}{2} \sum_{c=1}^2  \left\| \E [\vz_i | \hat{y}_i = c] - \E \vz_i \right\|_2^2 \\
        &= \frac{1}{2} \cdot \left( \frac{1}{(\E \hat{y}_i)^2} + \frac{1}{(1 - \E \hat{y}_i)^2} \right) \cdot \frac{1}{Z(\vw, s)^2} \cdot  \\
        &\quad\ \left( 4\Cov(\hat{y}_i, y)^2 \cdot \| \vmu \odot \vwcls \|_2^2 + \| \Cov(\hat{y}_i, \vvirr) \odot \vwirr \|_2^2 +  \| \Cov(\hat{y}_i, \vvnoise) \odot \vwnoise \|_2^2\right) \\
        &= \frac{1}{2} \cdot \left( \frac{1}{(\E \hat{y}_i)^2} + \frac{1}{(1 - \E \hat{y}_i)^2} \right) \cdot \frac{4 \sigmayhy^2 \cdot \| \vmu \odot \vwcls \|_2^2 + \| \vsigmairr \odot \vwirr \|_2^2 +  \| \vsigmanoise \odot \vwnoise \|_2^2 }{\| \vmu \odot \vwcls \|_2^2 + \| \vwirr \|_2^2 + s^2 \cdot \| \vdelta \odot \vwshift \|_2^2 + s^2 \cdot \| \vwnoise \|_2^2}
    \end{align*}
    As a simple correctness check, when the pseudo-label $\hat{y}_i = y_i, \forall i$, substituting $\E\hat{y}_i = \E y_i = \frac{1}{2}$, $\sigmayhy = \Var(y_i) = \frac{1}{4}$, $\vsigmairr = \vzero$, and $\vsigmanoise = \vzero$ recovers the result of GT-inter variance $\gVGTinter$ in Theorem \ref{thm:var}. 
    
    Finally, we compute the gradients w.r.t. four components of $\vw$ at initialization. Note that although the pseudo-labels $\hat{y}_i$ depend on the model parameters, this dependence involves an $\argmax$ operation and is thus non-differentiable. Therefore, during optimization, we treat the pseudo-labels as fixed constants and do not backpropagate through them. Let $C(\E \hat{y}_i) = \frac{1}{(\E \hat{y}_i)^2} + \frac{1}{(1 - \E \hat{y}_i)^2} $, 
    \begin{align*}
        \nabla_{\vwcls} \gVPLinter 
            &= \frac{C(\E \hat{y}_i)}{2} \cdot \frac{(4 \sigmayhy^2 \| \vwirr \|_2^2 - \| \vsigmairr \odot \vwirr \|_2^2) + (4 \sigmayhy^2 s^2 \| \vdelta \odot \vwshift \|_2^2) + (4 \sigmayhy^2 \| \vwnoise \|_2^2 - \| \vsigmanoise \odot \vwnoise \|_2^2)}{Z(\vw, s)^4} \cdot \\
            &\quad\ 2 \vmu^2 \odot \vwcls \\
            &= C(\E \hat{y}_i) \cdot \frac{(4 \sigmayhy^2 \dirr - \| \vsigmairr \|_2^2) + 4 \sigmayhy^2 s^2 \| \vdelta \|_2^2 + (4 \sigmayhy^2 \dnoise - \| \vsigmanoise \|_2^2)}{( \| \vmu \|_2^2 + \dirr + s^2 \cdot (\| \vdelta \|_2^2 +  \dnoise) )^2} \cdot \vmu^2 \\
        \nabla_{\vwshift} \gVPLinter
            &= \frac{C(\E \hat{y}_i)}{2} \cdot - \frac{4 \sigmayhy^2 \cdot \| \vmu \odot \vwcls \|_2^2 + \| \vsigmairr \odot \vwirr \|_2^2 +  \| \vsigmanoise \odot \vwnoise \|_2^2}{Z(\vw, s)^4} \cdot s^2 \cdot 2 \vdelta^2 \odot \vwshift \\
            &= - C(\E \hat{y}_i) \cdot \frac{\gVPLinter}{\| \vmu \|_2^2 + \dirr + s^2 \cdot (\| \vdelta \|_2^2 +  \dnoise) } \cdot s^2 \cdot \vdelta^2 
    \end{align*}

\end{proof}

\newpage
\section{Experiments}

\subsection{Effect of corruptions} \label{appendix:corruption}

\subsubsection{CIFAR-10-C}

\begin{figure}[h!]
    \centering
    \includegraphics[width=1.0\linewidth]{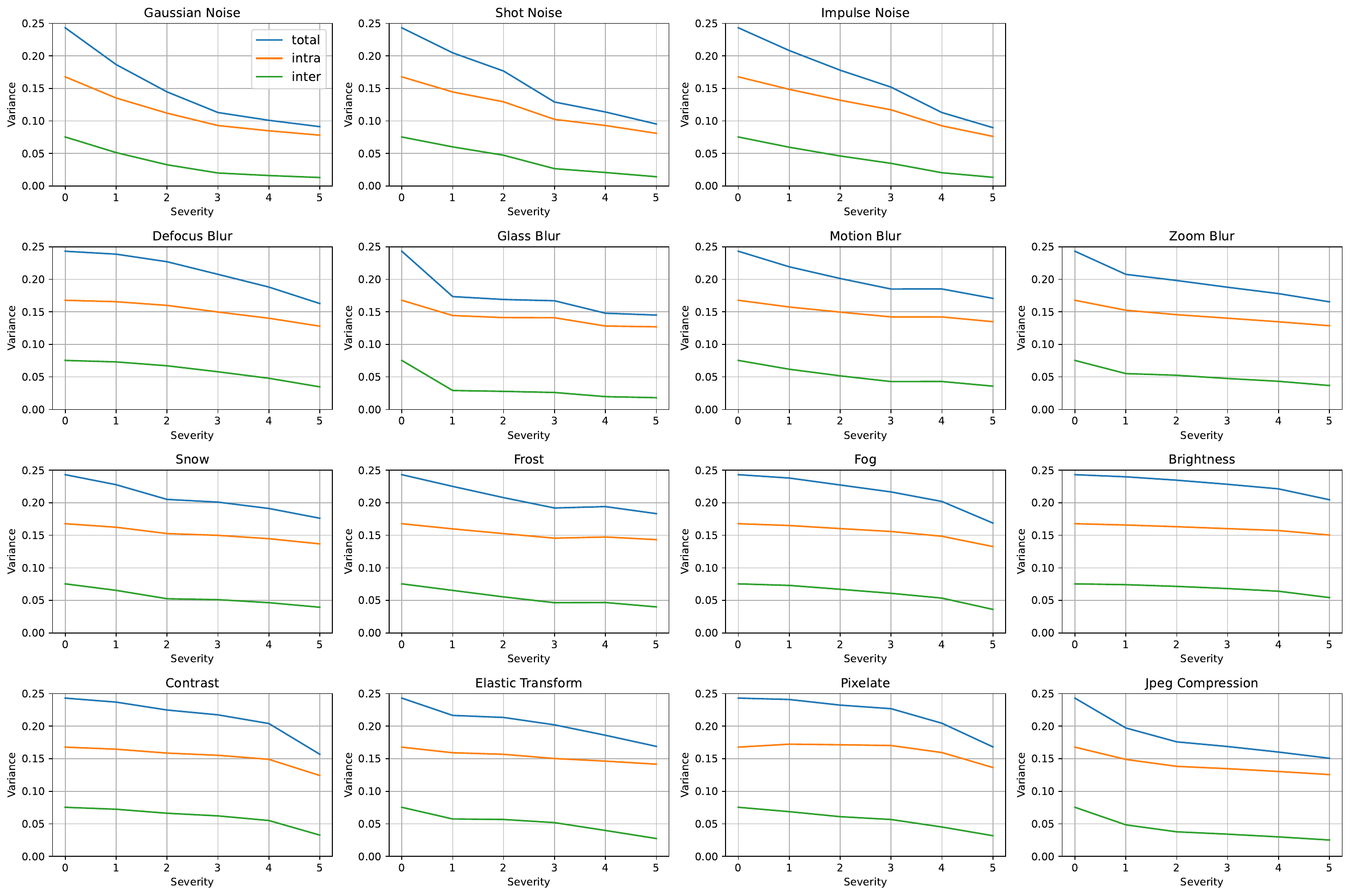}
    \caption{Effect of different levels of corruptions on ViT-B/32 on CIFAR-10-C. }
    \label{fig:vardec_cifar10}
\end{figure}

\begin{figure}[h!]
    \centering
    \includegraphics[width=0.75\linewidth]{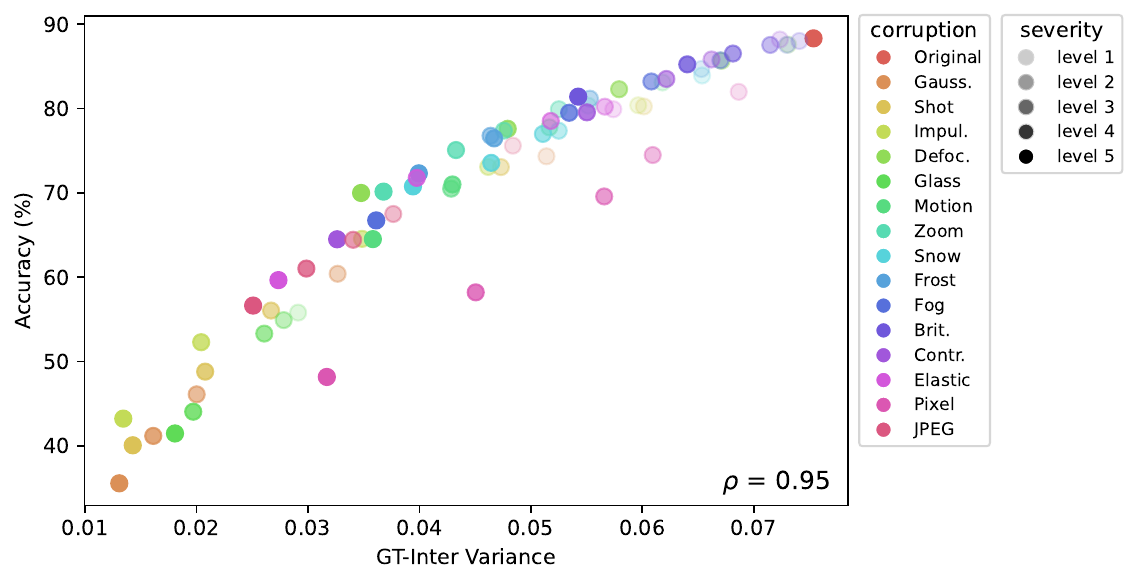}
    \caption{Correlation of GT-inter variance and classification accuracy of ViT-B/32 on CIFAR-10-C. }
    \label{fig:reg_cifar10}
\end{figure}

\begin{table}[h!]
    \centering
    \caption{Pearson correlation coefficients between accuracy and variances on ViT-B/32 on CIFAR-10-C. }
    \small{
    \begin{tabular}{cccc}
        \toprule
         & $\gVGTtotal$ & $\gVGTintra$ & $\gVGTinter$ \\
        \midrule
        Accuracy & 0.9104 & 0.8286 & \textbf{0.9483}  \\
        \bottomrule
    \end{tabular}
    }
\end{table}

\newpage

\subsubsection{CIFAR-100-C}

\begin{figure}[h!]
    \centering
    \includegraphics[width=1.0\linewidth]{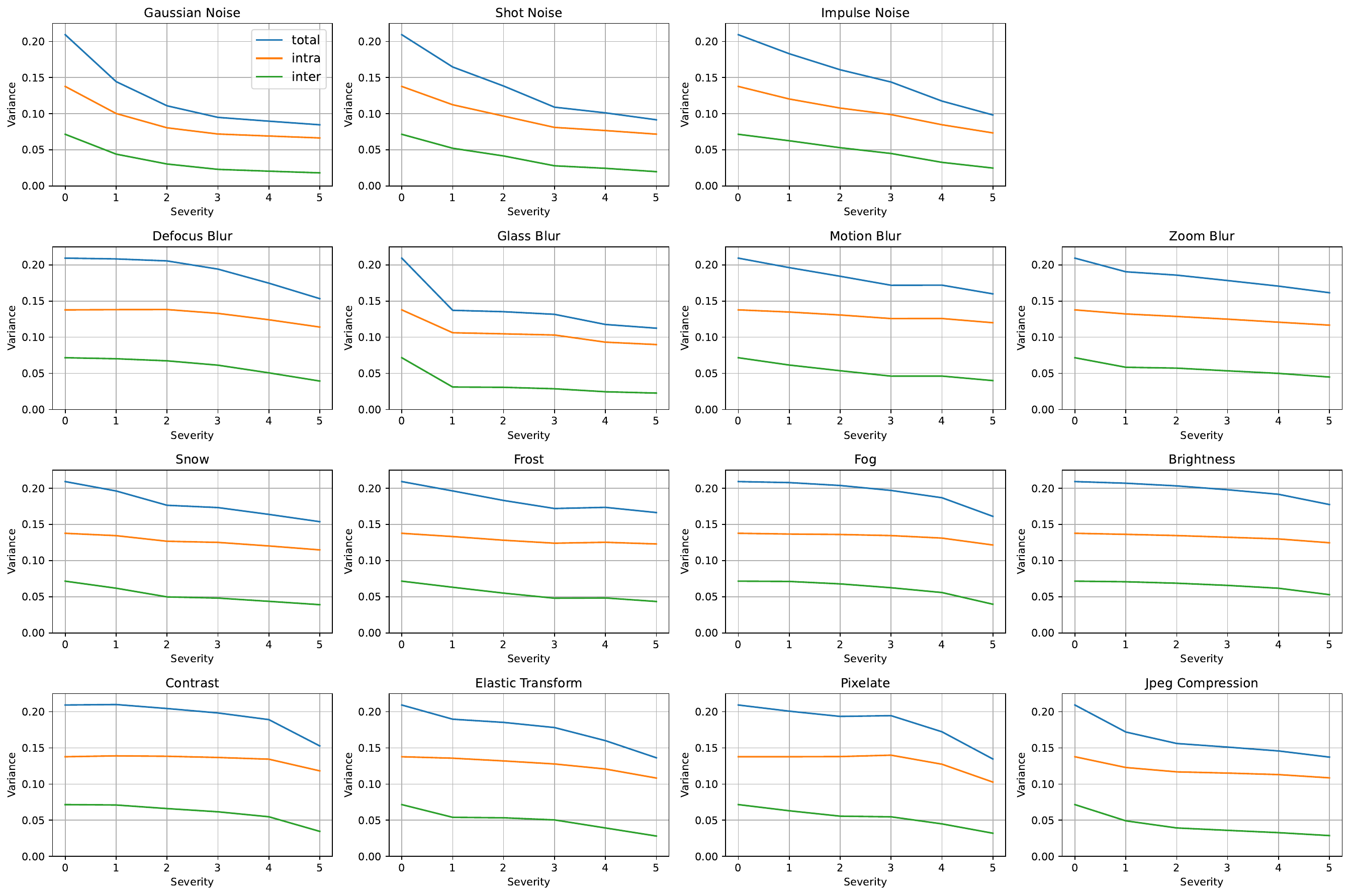}
    \caption{Effect of different levels of corruptions on ViT-B/16 on CIFAR-100-C. }
    \label{fig:vardec_cifar100}
\end{figure}

\begin{figure}[h!]
    \centering
    \includegraphics[width=0.75\linewidth]{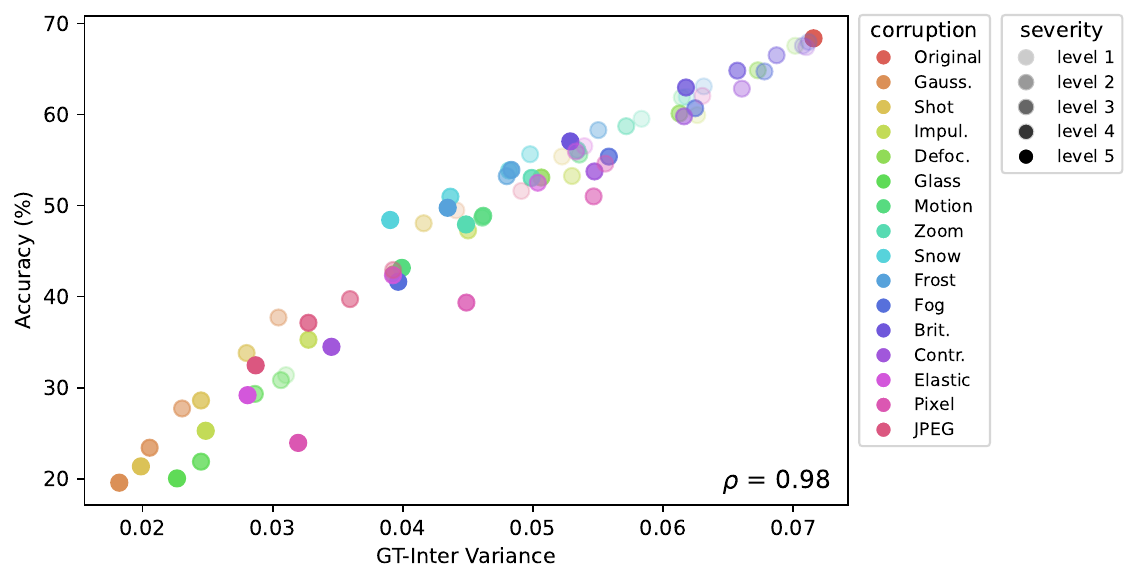}
    \caption{Correlation of GT-inter variance and classification accuracy of ViT-B/16 on CIFAR-100-C. }
    \label{fig:reg_cifar100}
\end{figure}

\begin{table}[h!]
    \centering
    \caption{Pearson correlation coefficients between accuracy and variances on ViT-B/16 on CIFAR-100-C. }
    \small{
    \begin{tabular}{cccc}
        \toprule
         & $\gVGTtotal$ & $\gVGTintra$ & $\gVGTinter$  \\
        \midrule
        Accuracy & 0.9364 & 0.8560 & \textbf{0.9752} \\
        \bottomrule
    \end{tabular}
    }
\end{table}

\newpage

\subsubsection{ImageNet-C}

\begin{figure}[h!]
    \centering
    \includegraphics[width=1.0\linewidth]{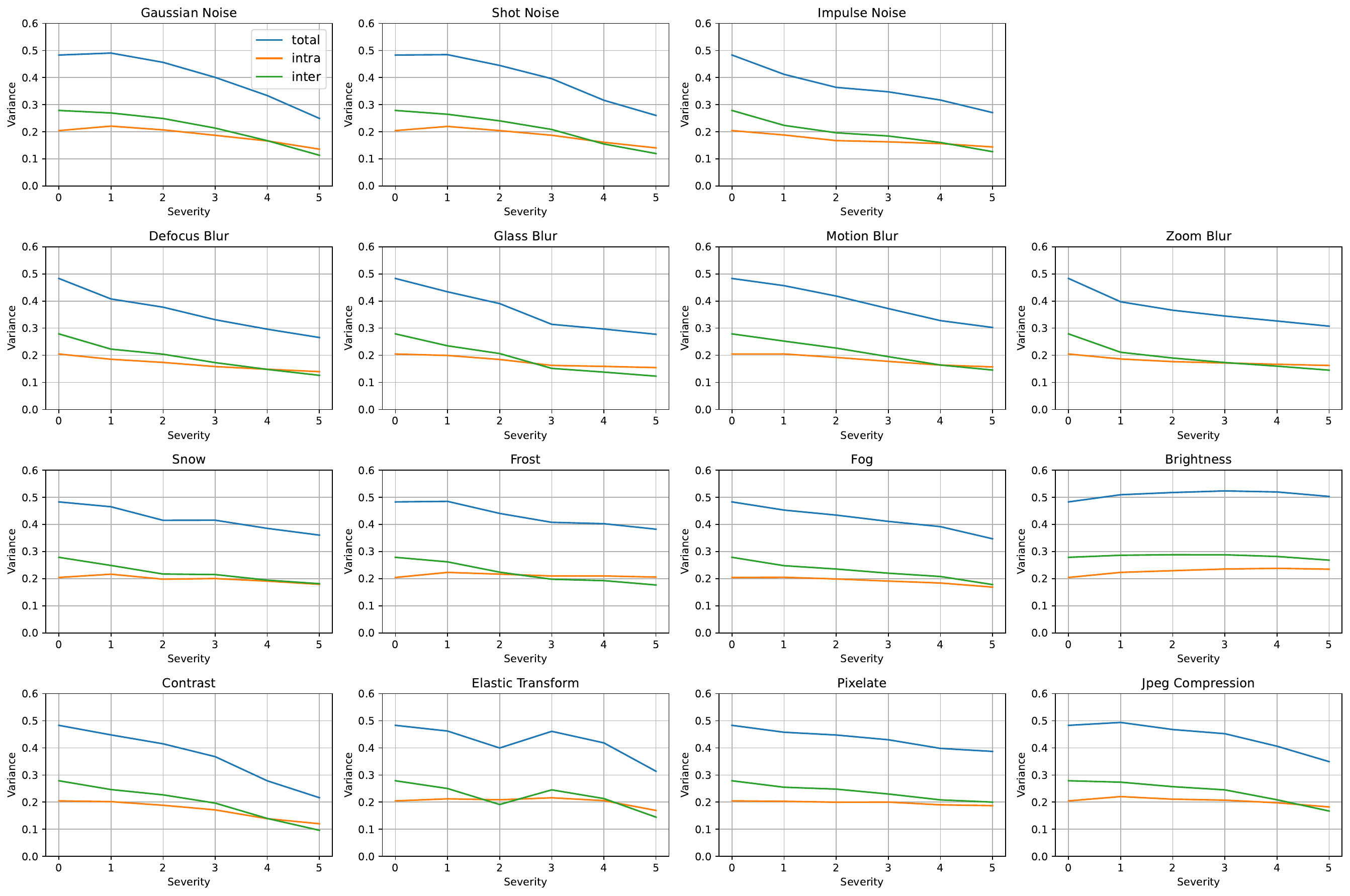}
    \caption{Effect of different levels of corruptions on ViT-L/14 on ImageNet-C. }
    \label{fig:vardec_imagenetc}
\end{figure}

\begin{figure}[h!]
    \centering
    \includegraphics[width=0.75\linewidth]{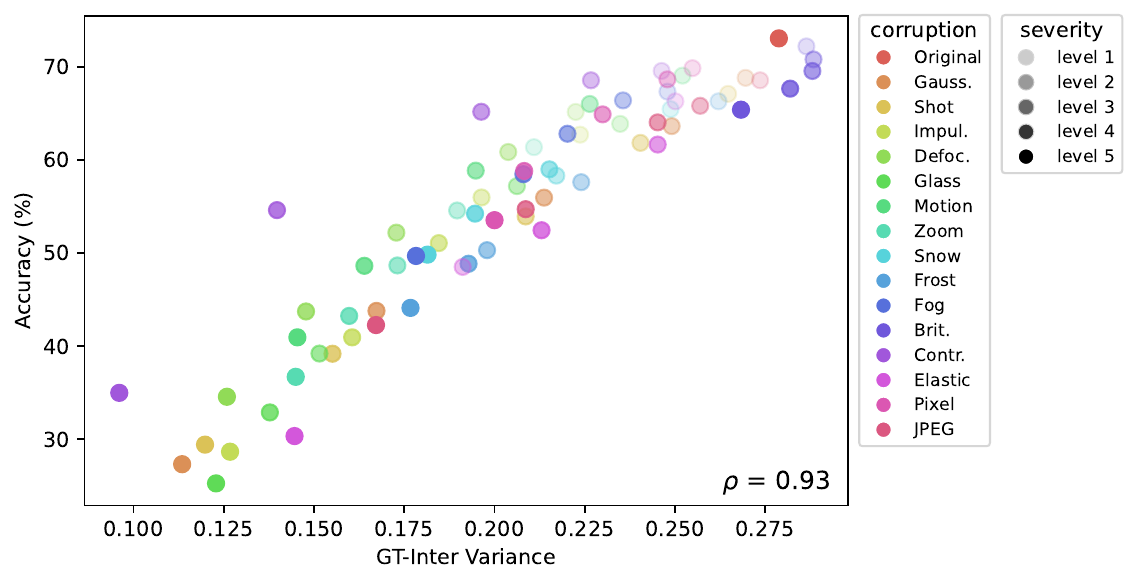}
    \caption{Correlation of GT-inter variance and classification accuracy of ViT-L/14 on ImageNet-C. }
    \label{fig:reg_imagenetc}
\end{figure}

\begin{table}[h!]
    \centering
    \caption{Pearson correlation coefficients between accuracy and variances on ViT-L/14 on ImageNet-C. }
    \small{
    \begin{tabular}{cccc}
        \toprule
         & $\gVGTtotal$ & $\gVGTintra$ & $\gVGTinter$  \\
        \midrule
        Accuracy & 0.8937 & 0.7753 & \textbf{0.9332}  \\
        \bottomrule
    \end{tabular}
    }
\end{table}

\newpage
\subsection{Experiment details} \label{appendix:exp:detail}

\subsubsection{Baselines}
For all methods, expect from TPT \cite{tpt} and CLIPArTT \cite{clipartt} which modifies the prompts, we use the 7 template in \cite{tip-adaptor}: 
\begin{itemize}
    \item ``\texttt{itap of a \{class\}}''
    \item ``\texttt{a bad photo of the \{class\}}''
    \item ``\texttt{a origami \{class\}}''
    \item ``\texttt{a photo of the large \{class\}}''
    \item ``\texttt{a \{class\} in a video game}''  
    \item ``\texttt{art of the \{class\}}''
    \item ``\texttt{a photo of the small \{class\}}''
\end{itemize}
The text embedding for each class $y$ is computed by
\begin{align*}
    \vt_y = \normalize\left( \sum_{\kappa = 1}^k \vt_{y, \kappa} \right), \text{where } \vt_{y, \kappa} = \normalize(\mathrm{text\_encoder(\{template_\kappa, classname_{\mathit{y}}\})})
\end{align*}

The following is our detailed handling method for other baselines and the usage of hyperparameters. The above hyperparameters are derived from those used in experiments reported in previous papers.

\begin{itemize}
    \item For all augmentation-based baselines (TPT \cite{tpt}, TPS \cite{tps}, Zero \cite{zero}, VTE \cite{vte}), we use AugMix to augment each test image 63 times to obtain a batch of 64 images, which includes the original image. We select 10\% of samples in the batch with lowest entropy to aggregate. 
    \item In TPT \cite{tpt}, the number of prompt tokens is 4, the prompt is initialized with ``a photo of a'', and class-specific contexts are disabled. We use the AdamW optimizer and adopt a learning rate of $0.005$, consistent with the setting used for ImageNet in the original papers. 
    \item In TDA \cite{tda}, positive cache is enabled with a shot capacity of 3, an adaptation strength ($\alpha$) of 2.0, and a sharpness ratio ($\beta$) of 5.0. The negative cache is enabled with a shot capacity of 2, an adaptation strength ($\alpha$) of 0.117, and a sharpness ratio ($\beta$) of 1.0, an entropy threshold between 0.2 and 0.5, and a mask threshold between 0.03 and 1.0.
    \item In DMN-ZS \cite{dmn}, the positive cache is enabled with a shot capacity of 50, an adaptation strength ($\alpha$) of 0.3, and a sharpness ratio ($\beta$) of 5.5.
    \item In TPS \cite{tps}, we also use the AdamW optimizer and adopt a learning rate of 0.005, consistent with the setting used for ImageNet in the original papers. 
    \item In WATT-S \cite{watt}, the learning rate is 0.001, the weight averaging is performed in a sequential manner, with 2 iterations per template and 5 total rounds of averaging.
    \item In CLIPArTT \cite{clipartt}, the learning rate is 0.001, the adaptation process runs for 10 steps, and the top 3 predicted classes are used to construct the pseudo-label prompt.
\end{itemize}

\subsubsection{Compute resources} \label{appendix:exp:resource}

All of our experiments are conducted on single NVIDIA Tesla V100 with 32GB memory, except for experiments on large batch size are conducted on single NVIDIA Tesla A100 with 80GB memory. 

\subsubsection{Licenses} \label{appendix:exp:license}

The corruption benchmark is licensed under the Apache-2.0 License, as indicated at \url{https://github.com/hendrycks/robustness}. CLIP is licensed under the MIT License, as stated at \url{https://github.com/openai/CLIP}.

\newpage
\subsection{Full results for RQ1} \label{appendix:exp:rq1}

\renewcommand{\meansd}[2]{#1{\scriptsize{ (#2)}}}

\begin{table}[h!]
    \centering
    \caption{Accuracy (mean (s.d.) \%) on corruption benchmarks with different batch sizes. }
    \label{tab:acc:full}
    \resizebox{1.0\linewidth}{!}{
    \setlength{\tabcolsep}{1.5mm}{
        \begin{tabular}{ccccccccccccccccc}
            \toprule
            & \multicolumn{16}{c}{ViT-B/32 on CIFAR-10-C} \\
            \cmidrule(lr){2-17}
            Batch Size & \multicolumn{3}{c}{Noise} & \multicolumn{4}{c}{Blur} & \multicolumn{4}{c}{Weather} & \multicolumn{4}{c}{Digital} & \multirow{2.5}{*}{Avg.} \\
            \cmidrule(lr){2-4} \cmidrule(lr){5-8} \cmidrule(lr){9-12} \cmidrule(lr){13-16}
            & Gauss. & Shot & Impul. & Defoc. & Glass & Motion & Zoom & Snow & Frost & Fog & Brit. & Contr. & Elastic & Pixel & JPEG & \\
            \midrule
            1 & \meansd{57.9}{0.5} & \meansd{61.3}{0.4} & \meansd{52.3}{0.3} & \meansd{76.2}{0.2} & \meansd{60.0}{0.2} & \meansd{76.9}{0.1} & \meansd{79.2}{0.2} & \meansd{79.1}{0.1} & \meansd{78.9}{0.3} & \meansd{75.2}{0.2} & \meansd{86.4}{0.1} & \meansd{76.6}{0.1} & \meansd{70.3}{0.2} & \meansd{63.4}{0.8} & \meansd{63.3}{0.3} & \meansd{70.5}{0.1} \\
            2 & \meansd{58.0}{0.5} & \meansd{61.3}{0.3} & \meansd{52.3}{0.4} & \meansd{76.2}{0.3} & \meansd{60.0}{0.2} & \meansd{76.9}{0.1} & \meansd{79.3}{0.2} & \meansd{79.1}{0.1} & \meansd{78.9}{0.3} & \meansd{75.2}{0.2} & \meansd{86.4}{0.1} & \meansd{76.6}{0.1} & \meansd{70.3}{0.2} & \meansd{63.4}{0.8} & \meansd{63.2}{0.2} & \meansd{70.5}{0.1} \\
            5 & \meansd{60.6}{0.1} & \meansd{62.9}{0.2} & \meansd{53.3}{0.3} & \meansd{76.2}{0.2} & \meansd{60.8}{0.2} & \meansd{77.0}{0.2} & \meansd{79.2}{0.2} & \meansd{79.1}{0.1} & \meansd{78.9}{0.3} & \meansd{75.1}{0.2} & \meansd{86.4}{0.1} & \meansd{76.8}{0.1} & \meansd{70.4}{0.3} & \meansd{64.6}{0.5} & \meansd{63.2}{0.1} & \meansd{71.0}{0.0} \\
            10 & \meansd{59.8}{0.4} & \meansd{62.8}{0.3} & \meansd{54.0}{0.3} & \meansd{76.0}{0.3} & \meansd{61.3}{0.3} & \meansd{77.1}{0.2} & \meansd{79.1}{0.3} & \meansd{79.0}{0.2} & \meansd{78.8}{0.3} & \meansd{75.1}{0.1} & \meansd{86.4}{0.2} & \meansd{76.8}{0.1} & \meansd{70.3}{0.2} & \meansd{65.7}{0.5} & \meansd{63.2}{0.1} & \meansd{71.0}{0.1} \\
            50 & \meansd{59.0}{0.5} & \meansd{62.4}{0.4} & \meansd{54.2}{0.3} & \meansd{75.8}{0.2} & \meansd{61.8}{0.3} & \meansd{77.1}{0.2} & \meansd{78.9}{0.2} & \meansd{79.0}{0.1} & \meansd{78.9}{0.2} & \meansd{75.2}{0.1} & \meansd{86.3}{0.1} & \meansd{76.9}{0.1} & \meansd{70.1}{0.3} & \meansd{66.6}{0.3} & \meansd{63.4}{0.2} & \meansd{71.0}{0.1} \\
            100 & \meansd{58.8}{0.7} & \meansd{62.0}{0.4} & \meansd{54.2}{0.3} & \meansd{75.6}{0.3} & \meansd{62.4}{0.4} & \meansd{77.0}{0.2} & \meansd{78.7}{0.3} & \meansd{78.9}{0.1} & \meansd{78.7}{0.3} & \meansd{75.1}{0.2} & \meansd{86.4}{0.1} & \meansd{77.1}{0.1} & \meansd{69.7}{0.2} & \meansd{67.2}{0.2} & \meansd{63.4}{0.2} & \meansd{71.0}{0.1} \\
            200 & \meansd{58.0}{1.0} & \meansd{61.3}{0.5} & \meansd{54.7}{0.5} & \meansd{75.0}{0.2} & \meansd{62.2}{0.6} & \meansd{77.0}{0.2} & \meansd{78.4}{0.3} & \meansd{78.5}{0.2} & \meansd{78.6}{0.3} & \meansd{74.9}{0.1} & \meansd{86.2}{0.1} & \meansd{77.0}{0.1} & \meansd{68.0}{0.3} & \meansd{67.0}{0.5} & \meansd{62.0}{0.2} & \meansd{70.6}{0.1} \\
            % 500 & \meansd{57.4}{0.7} & \meansd{61.2}{0.5} & \meansd{54.9}{0.6} & \meansd{74.4}{0.2} & \meansd{61.2}{0.7} & \meansd{77.0}{0.1} & \meansd{77.9}{0.2} & \meansd{78.0}{0.1} & \meansd{77.3}{0.2} & \meansd{74.7}{0.2} & \meansd{86.1}{0.1} & \meansd{76.9}{0.1} & \meansd{65.8}{0.3} & \meansd{66.7}{0.8} & \meansd{60.4}{0.4} & \meansd{70.0}{0.1} \\
            \midrule
            & \multicolumn{16}{c}{ViT-B/16 on CIFAR-100-C} \\
            \cmidrule(lr){2-17}
            Batch Size & \multicolumn{3}{c}{Noise} & \multicolumn{4}{c}{Blur} & \multicolumn{4}{c}{Weather} & \multicolumn{4}{c}{Digital} & \multirow{2.5}{*}{Avg.} \\
            \cmidrule(lr){2-4} \cmidrule(lr){5-8} \cmidrule(lr){9-12} \cmidrule(lr){13-16}
            & Gauss. & Shot & Impul. & Defoc. & Glass & Motion & Zoom & Snow & Frost & Fog & Brit. & Contr. & Elastic & Pixel & JPEG & \\
            \midrule
            1 & \meansd{23.9}{0.4} & \meansd{26.1}{0.4} & \meansd{37.2}{0.2} & \meansd{50.5}{0.1} & \meansd{27.0}{0.2} & \meansd{49.6}{0.3} & \meansd{55.3}{0.2} & \meansd{53.0}{0.2} & \meansd{51.6}{0.2} & \meansd{50.1}{0.2} & \meansd{65.5}{0.1} & \meansd{46.6}{0.3} & \meansd{36.7}{0.2} & \meansd{34.4}{0.5} & \meansd{38.7}{0.6} & \meansd{43.1}{0.1} \\
            2 & \meansd{23.9}{0.4} & \meansd{26.2}{0.4} & \meansd{37.2}{0.2} & \meansd{50.5}{0.1} & \meansd{27.0}{0.2} & \meansd{49.7}{0.3} & \meansd{55.3}{0.2} & \meansd{53.0}{0.2} & \meansd{51.6}{0.1} & \meansd{50.1}{0.2} & \meansd{65.5}{0.2} & \meansd{46.7}{0.3} & \meansd{36.7}{0.2} & \meansd{34.5}{0.5} & \meansd{38.7}{0.6} & \meansd{43.1}{0.1} \\
            5 & \meansd{24.8}{0.3} & \meansd{27.1}{0.6} & \meansd{37.5}{0.2} & \meansd{50.6}{0.1} & \meansd{27.1}{0.2} & \meansd{49.7}{0.3} & \meansd{55.4}{0.1} & \meansd{53.0}{0.2} & \meansd{51.7}{0.2} & \meansd{50.3}{0.2} & \meansd{65.5}{0.2} & \meansd{47.0}{0.3} & \meansd{36.7}{0.2} & \meansd{34.4}{0.6} & \meansd{38.7}{0.5} & \meansd{43.3}{0.1} \\
            10 & \meansd{26.4}{0.7} & \meansd{28.9}{0.8} & \meansd{38.0}{0.2} & \meansd{50.6}{0.1} & \meansd{27.1}{0.2} & \meansd{49.7}{0.2} & \meansd{55.4}{0.1} & \meansd{53.0}{0.2} & \meansd{51.7}{0.2} & \meansd{50.4}{0.1} & \meansd{65.5}{0.1} & \meansd{47.5}{0.3} & \meansd{36.8}{0.2} & \meansd{34.4}{0.6} & \meansd{38.7}{0.6} & \meansd{43.6}{0.1} \\
            20 & \meansd{29.4}{0.5} & \meansd{30.8}{0.7} & \meansd{38.6}{0.2} & \meansd{50.7}{0.2} & \meansd{27.1}{0.2} & \meansd{49.9}{0.2} & \meansd{55.5}{0.1} & \meansd{53.0}{0.2} & \meansd{51.8}{0.1} & \meansd{50.6}{0.2} & \meansd{65.6}{0.1} & \meansd{48.1}{0.2} & \meansd{36.8}{0.2} & \meansd{34.4}{0.7} & \meansd{38.7}{0.5} & \meansd{44.1}{0.1} \\
            50 & \meansd{31.4}{0.3} & \meansd{33.0}{0.5} & \meansd{39.4}{0.3} & \meansd{50.9}{0.1} & \meansd{26.8}{0.2} & \meansd{50.0}{0.3} & \meansd{55.5}{0.2} & \meansd{53.1}{0.2} & \meansd{51.9}{0.1} & \meansd{50.8}{0.1} & \meansd{65.8}{0.2} & \meansd{49.2}{0.1} & \meansd{36.9}{0.2} & \meansd{34.6}{1.0} & \meansd{38.3}{0.5} & \meansd{44.5}{0.1} \\
            100 & \meansd{31.1}{0.2} & \meansd{33.4}{0.4} & \meansd{40.0}{0.3} & \meansd{51.0}{0.2} & \meansd{27.1}{0.3} & \meansd{50.1}{0.3} & \meansd{55.5}{0.1} & \meansd{53.1}{0.2} & \meansd{51.9}{0.1} & \meansd{51.0}{0.1} & \meansd{65.8}{0.1} & \meansd{49.5}{0.2} & \meansd{36.7}{0.2} & \meansd{34.6}{0.9} & \meansd{37.4}{0.4} & \meansd{44.5}{0.1} \\
            200 & \meansd{30.8}{0.4} & \meansd{33.5}{0.5} & \meansd{40.2}{0.2} & \meansd{51.2}{0.2} & \meansd{27.5}{0.5} & \meansd{50.3}{0.2} & \meansd{55.6}{0.2} & \meansd{53.3}{0.2} & \meansd{52.0}{0.2} & \meansd{51.2}{0.1} & \meansd{65.9}{0.3} & \meansd{49.6}{0.2} & \meansd{36.8}{0.1} & \meansd{34.9}{0.7} & \meansd{36.8}{0.3} & \meansd{44.6}{0.1} \\
            % 500 & \meansd{30.1}{0.4} & \meansd{32.9}{0.9} & \meansd{40.0}{0.2} & \meansd{51.1}{0.1} & \meansd{28.1}{0.7} & \meansd{49.9}{0.3} & \meansd{55.0}{0.2} & \meansd{53.2}{0.3} & \meansd{51.9}{0.3} & \meansd{50.9}{0.3} & \meansd{65.5}{0.2} & \meansd{49.3}{0.4} & \meansd{36.5}{0.1} & \meansd{34.5}{0.6} & \meansd{35.2}{0.2} & \meansd{44.3}{0.1} \\

            \midrule
            & \multicolumn{16}{c}{ViT-L/14 on ImageNet-C} \\
             \cmidrule(lr){2-17}
            Batch Size & \multicolumn{3}{c}{Noise} & \multicolumn{4}{c}{Blur} & \multicolumn{4}{c}{Weather} & \multicolumn{4}{c}{Digital} & \multirow{2.5}{*}{Avg.} \\
            \cmidrule(lr){2-4} \cmidrule(lr){5-8} \cmidrule(lr){9-12} \cmidrule(lr){13-16}
            & Gauss. & Shot & Impul. & Defoc. & Glass & Motion & Zoom & Snow & Frost & Fog & Brit. & Contr. & Elastic & Pixel & JPEG & \\
            \midrule
            1 & \meansd{31.6}{0.5} & \meansd{33.2}{0.1} & \meansd{36.3}{0.2} & \meansd{41.0}{0.2} & \meansd{37.6}{0.3} & \meansd{46.5}{0.4} & \meansd{41.3}{0.4} & \meansd{54.0}{0.1} & \meansd{43.4}{0.1} & \meansd{57.4}{0.3} & \meansd{67.7}{0.1} & \meansd{45.7}{0.3} & \meansd{41.5}{0.1} & \meansd{55.7}{0.5} & \meansd{54.8}{0.2} & \meansd{45.8}{0.1} \\
            2 & \meansd{31.9}{0.4} & \meansd{33.0}{0.3} & \meansd{37.2}{0.2} & \meansd{40.1}{0.2} & \meansd{37.6}{0.2} & \meansd{46.7}{0.3} & \meansd{42.1}{0.4} & \meansd{54.5}{0.3} & \meansd{43.8}{0.2} & \meansd{57.7}{0.2} & \meansd{67.7}{0.1} & \meansd{47.7}{0.5} & \meansd{41.5}{0.3} & \meansd{57.1}{0.3} & \meansd{55.1}{0.1} & \meansd{46.2}{0.1} \\
            5 & \meansd{32.4}{0.4} & \meansd{33.4}{0.3} & \meansd{37.1}{0.3} & \meansd{39.7}{0.4} & \meansd{37.6}{0.3} & \meansd{46.7}{0.5} & \meansd{43.4}{0.3} & \meansd{55.6}{0.4} & \meansd{44.4}{0.4} & \meansd{57.7}{0.2} & \meansd{67.8}{0.1} & \meansd{49.4}{0.7} & \meansd{42.0}{0.4} & \meansd{57.8}{0.2} & \meansd{55.3}{0.1} & \meansd{46.7}{0.1} \\
            10 & \meansd{32.6}{0.3} & \meansd{33.6}{0.2} & \meansd{36.8}{0.1} & \meansd{39.7}{0.4} & \meansd{37.6}{0.5} & \meansd{46.8}{0.4} & \meansd{44.1}{0.4} & \meansd{55.3}{0.2} & \meansd{45.2}{0.7} & \meansd{57.5}{0.2} & \meansd{67.7}{0.1} & \meansd{49.4}{0.8} & \meansd{42.8}{0.3} & \meansd{58.3}{0.1} & \meansd{55.2}{0.3} & \meansd{46.8}{0.1} \\
            20 & \meansd{33.0}{0.3} & \meansd{34.3}{0.3} & \meansd{37.3}{0.2} & \meansd{39.6}{0.4} & \meansd{37.2}{0.4} & \meansd{46.6}{0.3} & \meansd{45.1}{0.5} & \meansd{55.2}{0.1} & \meansd{46.6}{0.7} & \meansd{57.5}{0.1} & \meansd{67.7}{0.2} & \meansd{48.9}{0.8} & \meansd{43.9}{0.4} & \meansd{58.2}{0.2} & \meansd{54.6}{0.7} & \meansd{47.0}{0.2} \\
            50 & \meansd{33.2}{0.4} & \meansd{35.1}{0.2} & \meansd{37.6}{0.5} & \meansd{38.8}{0.3} & \meansd{36.9}{0.4} & \meansd{47.1}{0.2} & \meansd{45.4}{0.6} & \meansd{55.0}{0.3} & \meansd{48.3}{0.7} & \meansd{57.5}{0.2} & \meansd{67.4}{0.2} & \meansd{47.0}{1.1} & \meansd{45.1}{0.1} & \meansd{57.9}{0.3} & \meansd{54.8}{0.6} & \meansd{47.1}{0.1} \\
            100 & \meansd{33.8}{0.5} & \meansd{34.9}{0.3} & \meansd{37.6}{0.2} & \meansd{38.6}{0.4} & \meansd{37.2}{0.5} & \meansd{47.4}{0.5} & \meansd{45.7}{0.4} & \meansd{55.0}{0.3} & \meansd{48.9}{0.2} & \meansd{57.5}{0.4} & \meansd{67.2}{0.2} & \meansd{44.2}{1.4} & \meansd{46.0}{0.3} & \meansd{57.4}{0.2} & \meansd{54.0}{0.4} & \meansd{47.0}{0.2} \\
            200 & \meansd{33.9}{0.2} & \meansd{34.7}{0.5} & \meansd{37.7}{0.1} & \meansd{38.7}{0.1} & \meansd{37.2}{0.5} & \meansd{47.2}{0.4} & \meansd{45.3}{0.5} & \meansd{54.7}{0.3} & \meansd{49.1}{0.2} & \meansd{57.5}{0.3} & \meansd{67.3}{0.2} & \meansd{42.0}{1.9} & \meansd{46.3}{0.5} & \meansd{57.0}{0.3} & \meansd{53.8}{0.7} & \meansd{46.8}{0.1} \\
            \bottomrule
        \end{tabular}
    }
    }
\end{table}

\renewcommand{\meansd}[2]{#1 $\pm$ #2}

\subsection{Full results for RQ2} \label{appendix:exp:rq2}

\begin{figure}[h!]
    \centering
    \includegraphics[width=1.0\linewidth]{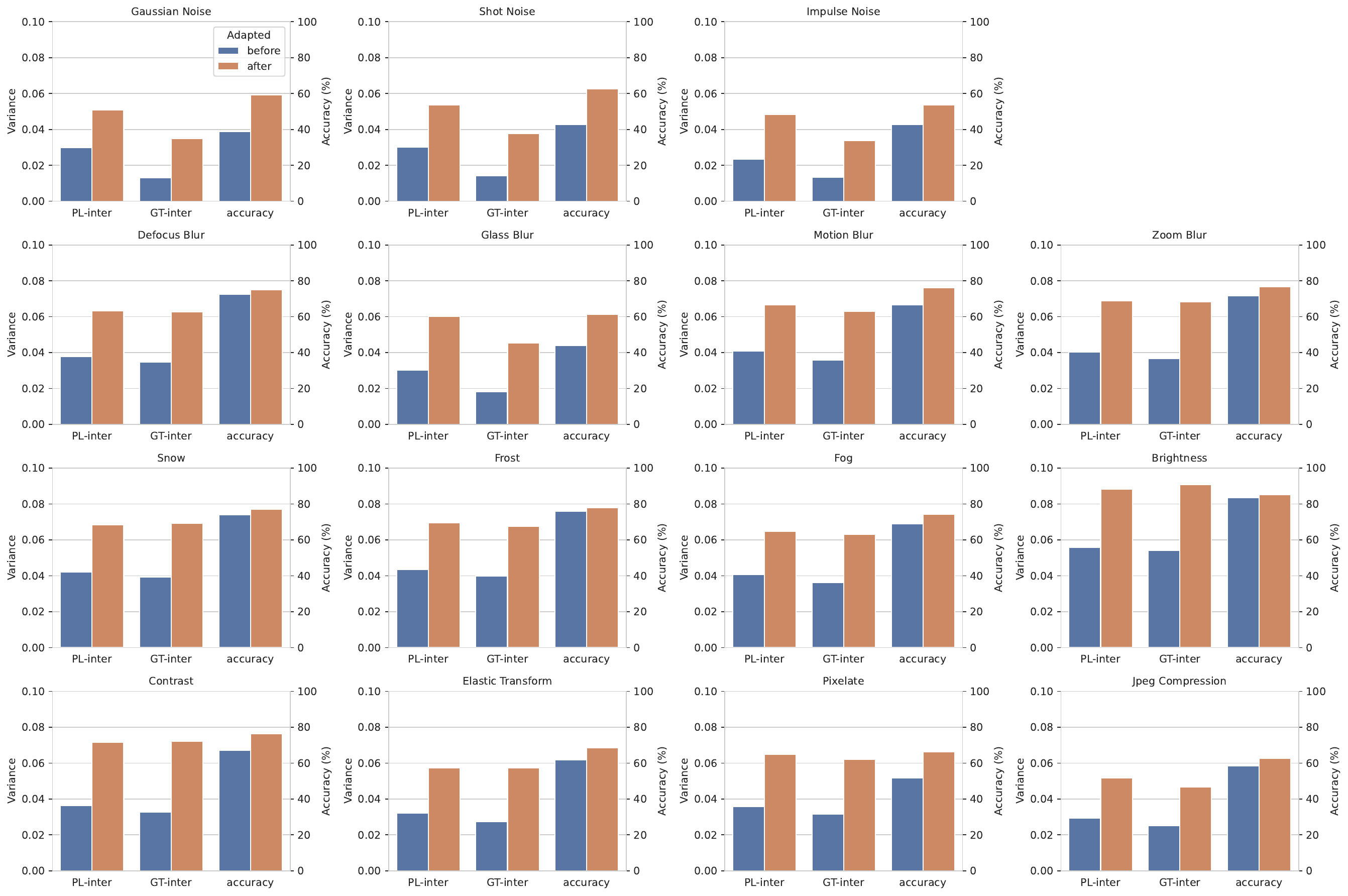}
    \caption{Variance collapse mitigation on CIFAR-10-C.}
    \label{fig:varmax_cifar10c}
\end{figure}
 
\begin{figure}[h!]
    \centering
    \includegraphics[width=1.0\linewidth]{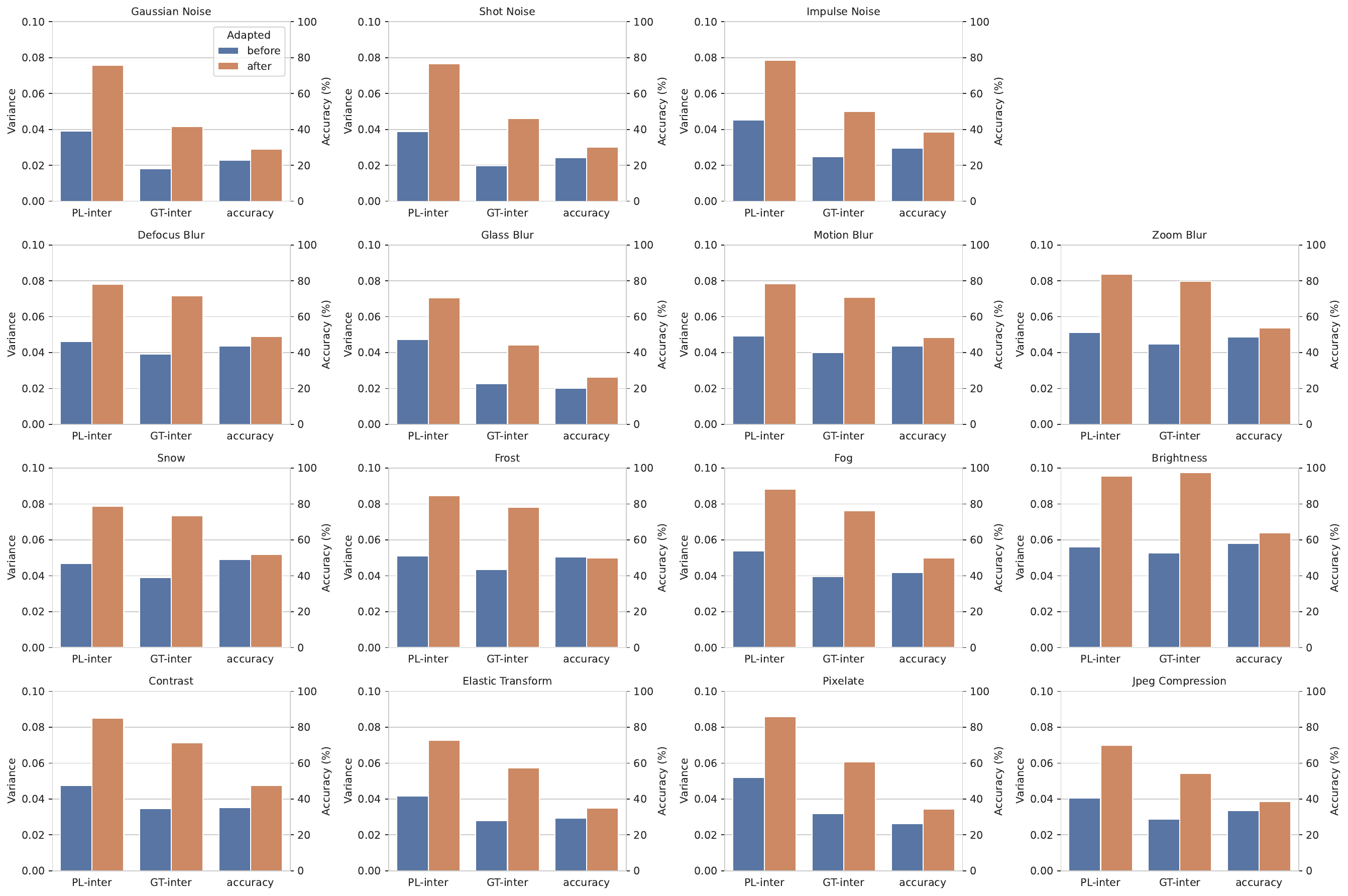}
    \caption{Variance collapse mitigation on CIFAR-100-C.}
    \label{fig:varmax_cifar100c}
\end{figure}

\begin{figure}[h!]
    \centering
    \includegraphics[width=1.0\linewidth]{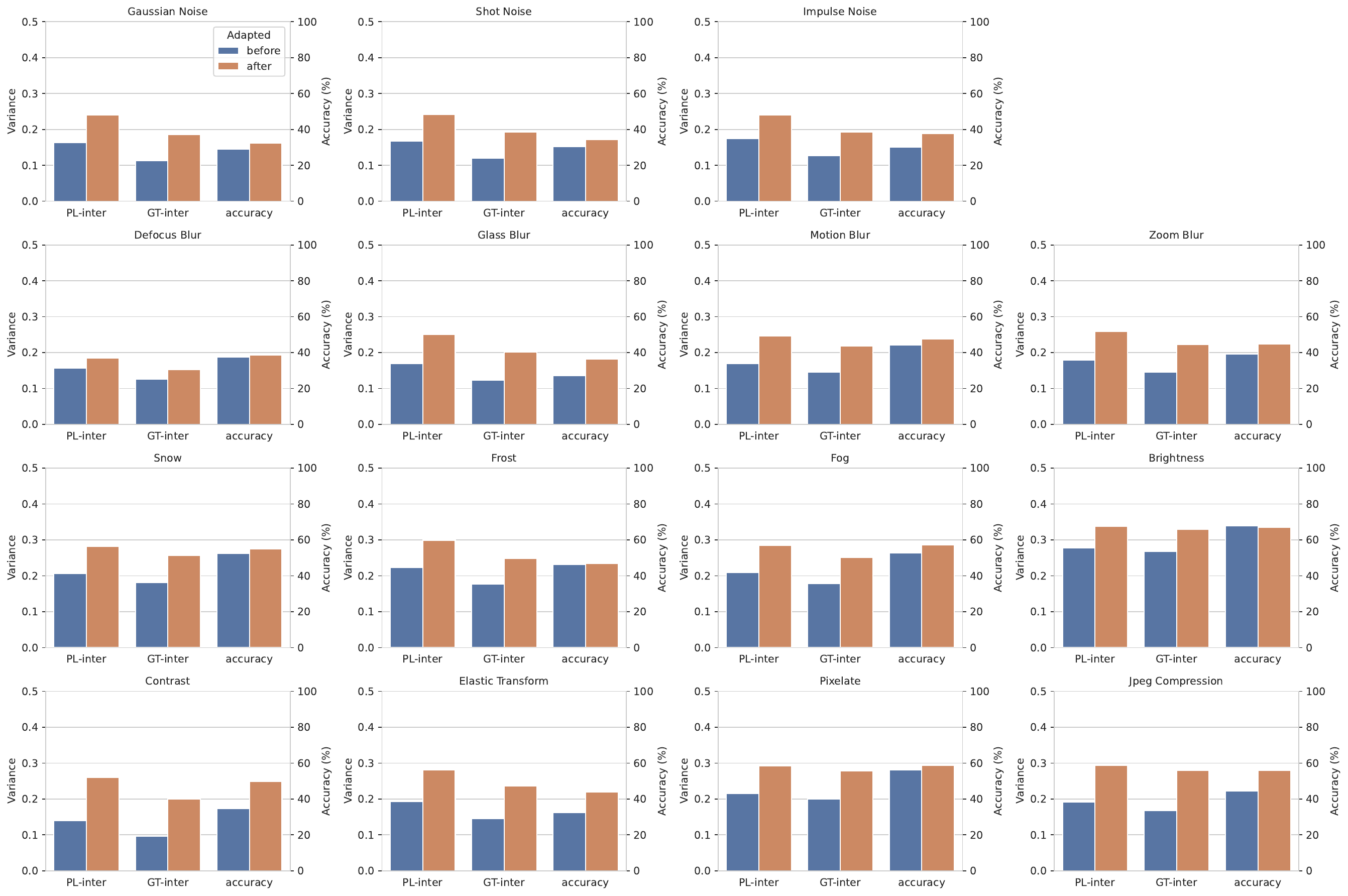}
    \caption{Variance collapse mitigation on ImageNet-C.}
    \label{fig:varmax_imagenetc}
\end{figure}

\newpage
\subsection{Ablation on layers to adapt}
\label{appendix:exp:layer}

{\algname} adapts all \emph{LayerNorm} layers in the vision encoder. Alternative choices include updating only the last block, the last MLP layer, or the first patching layer. Our comparison shows that updating \emph{all LayerNorm} layers yields the best performance. For further discussions on which layers are most effective to adapt, we refer readers to related studies \cite{surgical,atp}. 

\begin{table}[H]
    \centering
    \caption{Comparison of different parts of the image encoder to update at test-time on CIFAR-10-C with ViT-B/32.}
    \label{tab:ablate_layers}
    \small
    \begin{tabular}{lc}
        \toprule
        Layers to adapt   & Accuracy (\%) \\
        \midrule
        Last block           & 63.4 \\
        Last MLP             & 62.9 \\
        Patching layer       & 63.6 \\
        All LayerNorm ({\algname}) & \textbf{71.0} \\
        \bottomrule
    \end{tabular}
\end{table}

\subsection{Hyperparameter sensitivity} \label{appendix:exp:hp}

In this subsection, we provide the results of hyperparameter sensitivity experiments on CIFAR-10-C and CIFAR-100-C. 

\begin{figure}[h!]
    \centering
    \includegraphics[width=0.7\linewidth]{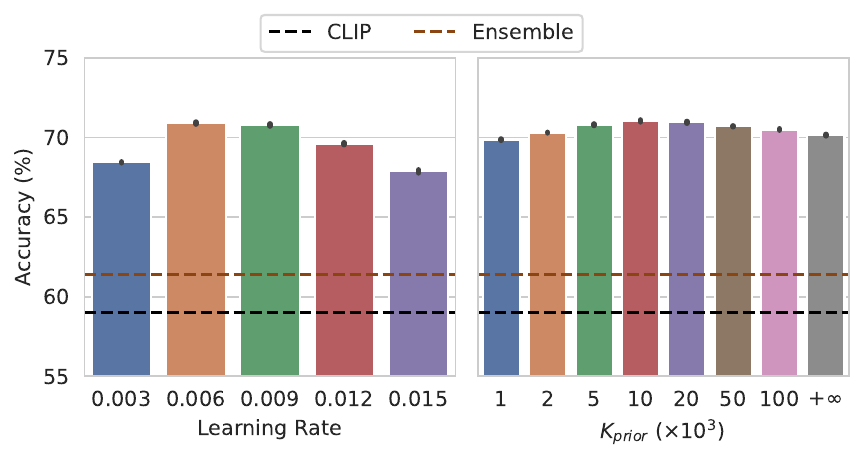}
    \caption{Hyperparameter sensitivity on CIFAR-10-C. }
        \label{fig:hp_sensitivity_cifar10c}
\end{figure}

\begin{figure}[h!]
    \centering
    \includegraphics[width=0.7\linewidth]{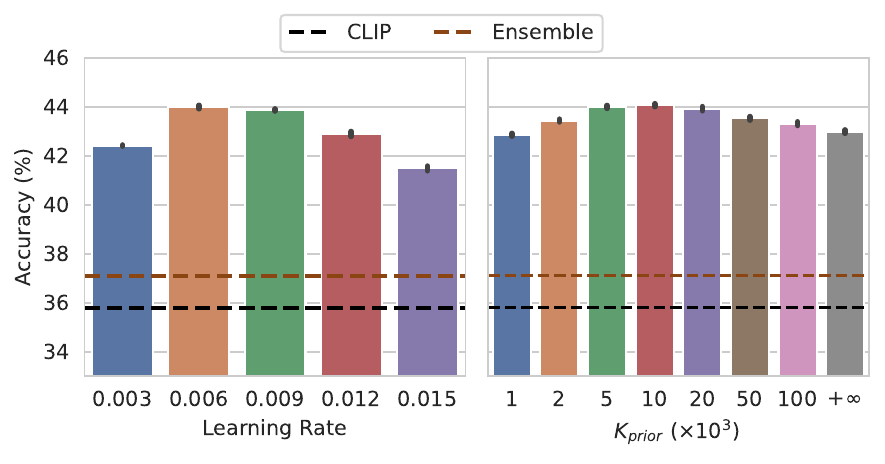}
    \caption{Hyperparameter sensitivity on CIFAR-100-C. }
        \label{fig:hp_sensitivity_cifar100c}
\end{figure}

% \begin{figure}[h!]
%     \centering
%     \includegraphics[width=0.7\linewidth]{figure/hyperparameter/hp_imagenetc.pdf}
%     \caption{Hyperparameter sensitivity on ImageNet-C. }
%         \label{fig:hp_sensitivity_imagenetc}
% \end{figure}

\newpage
\subsection{Experiments on clean datasets}
\label{appendix:exp:clean}

In this subsection, we offer the comparison between {\algname} and currently outstanding baselines, CLIPArTT \cite{clipartt}, WATT-S \cite{watt}, TDA \cite{tda}, DMN-ZS \cite{dmn}, Tent \cite{tent}, and ETA \cite{eata}, on clean (non-corruption) datasets with ViT-B/32 on CIFAR-10, ViT-B/16 on CIFAR-100, and ViT-L/14 on ImageNet.

\begin{table}[H]
    \centering
    \caption{Accuracy (\%) on clean datasets.}
    \label{tab:clean}
    \small
    \begin{tabular}{lccc}
        \toprule
        Method & ViT-B/32 on CIFAR-10 & ViT-B/16 on CIFAR-100 & ViT-L/14 on ImageNet \\
        \midrule
        CLIP      & 88.3 & 68.4 & 73.0 \\
        CLIPArTT  & 89.1 & 70.2 & 72.1 \\
        WATT-S    & 89.8 & 72.3 & 74.5 \\
        TDA       & 89.6 & 70.1 & 73.4 \\
        DMN-ZS    & 90.2 & 69.4 & 73.1 \\
        Tent      & 91.1 & 72.2 & 73.4 \\
        ETA       & 91.4 & 73.0 & 73.6 \\
        {\algname}      & \textbf{91.6} & \textbf{74.1} & \textbf{75.6} \\
        \bottomrule
    \end{tabular}
\end{table}

\subsection{Experiments on ImageNet variants}
\label{appendix:exp:imagenet_variants}

In this subsection, we provide the comparison between {\algname} and currently outstanding baselines, CLIPArTT \cite{clipartt}, WATT-S \cite{watt}, TDA \cite{tda}, DMN-ZS \cite{dmn}, Tent \cite{tent}, and ETA \cite{eata} in ImageNet variants (-A, -V2, -R, -Sketch) datasets with ViT-B/16.

\begin{table}[H] % requires \usepackage{float}
    \centering
    \caption{Accuracy (\%) on ImageNet variants with ViT-B/16.}
    \label{tab:other_imagenet}
    \small
    \begin{tabular}{lcccc}
        \toprule
        Method  & ImageNet-A & ImageNet-V2 & ImageNet-R & ImageNet-Sketch \\
        \midrule
        CLIP     & 49.2 & 60.4 & 72.7 & 44.9 \\
        CLIPArTT & 49.6 & 60.5 & 72.8 & 45.0 \\
        WATT-S   & 51.7 & 61.2 & 75.7 & 47.0 \\
        TDA      & 51.0 & 61.2 & 73.9 & 46.4 \\
        DMN-ZS   & 49.7 & 60.5 & 73.0 & 45.4 \\
        Tent     & 51.9 & 61.0 & 77.0 & 45.4 \\
        ETA      & 52.0 & 61.0 & 77.4 & 46.8 \\
        {\algname}     & \textbf{54.7} & \textbf{62.6} & \textbf{78.1} & \textbf{48.4} \\
        \bottomrule
    \end{tabular}
\end{table}

\subsection{Mixture corruption datasets}
\label{appendix:exp:mixture}

In this subsection, we compare between {\algname} and currently outstanding baselines, CLIPArTT \cite{clipartt}, WATT-S \cite{watt}, TDA \cite{tda}, and DMN-ZS \cite{dmn}, on mixture of 15 types of corruption datasets on CIFAR-10-C with ViT-B/32, CIFAR-100-C with ViT-B/16, and ImageNet-C with ViT-L/14. While the results in the main text are obtained by testing on each corruption type separately, here we first mix the data from all 15 corruptions together and then perform evaluation on this mixed-domain setting, following the setup in \cite{sar}.

\begin{table}[H]
    \centering
    \caption{Accuracy on Mixture of 15 Types of Corruptions.}
    \label{tab:mix15}
    \small
    \begin{tabular}{lccc}
        \toprule
        Method   & CIFAR-10-C & CIFAR-100-C & ImageNet-C \\
        \midrule
        CLIP      & 59.0 & 35.8 & 39.6 \\
        TDA       & 62.1 & 38.3 & 42.3 \\
        DMN-ZS    & 60.2 & 36.0 & 39.9 \\
        WATT-S   & 63.6 & 39.0 & 43.9 \\
        CLIPArTT  & 56.9 & 38.7 & 40.5 \\
        {\algname}      & \textbf{65.9} & \textbf{39.8} & \textbf{45.2} \\
        \bottomrule
    \end{tabular}
\end{table}

%%%%%%%%%%%%%%%%%%%%%%%%%%%%%%%%%%%%%%%%%%%%%%%%%%%%%%%%%%%%

\end{document}